\documentclass[letterpaper, 10 pt, journal, twoside]{ieeetran}
\usepackage{cite}
\usepackage{amsmath,amssymb,amsfonts}
\usepackage{graphicx}
\usepackage{textcomp}
\usepackage{xcolor}
\usepackage{pifont}

\usepackage{irap_acronyms}
\usepackage{irap_acronyms2}
\usepackage{irap_math}
\usepackage{irap_SIunits}
\usepackage{irap_misc}
\makeatletter
\let\NAT@parse\undefined
\makeatother
\usepackage[numbers]{natbib}
\usepackage{subfigure}
\usepackage{multirow}
\usepackage{adjustbox}
\usepackage{lipsum}
\usepackage{mathtools}

\DeclareMathOperator*{\argmax}{argmax}

\usepackage{algorithm,algpseudocode}

\usepackage{bm}
\usepackage{booktabs}

\usepackage{tensor}
\usepackage{dirtytalk}
\begin{document}

% \title{SIO-Mapper: Lane-level HD Map Construction Framework using Satellite Image and OpenStreetMap}
\title{SIO-Mapper: A Framework for Lane-Level HD Map Construction Using Satellite Images and OpenStreetMap with No On-Site Visits}

\author{Younghun Cho${}^{1}$ and Jee-Hwan Ryu${}^{1*}$
    \thanks{
      $^{1}$Younghun Cho, and Jee-Hwan Ryu are affiliated with the Department of Civil and Environmental Engineering, KAIST, Daejeon 34141, Korea {\tt\small \{dudgnsrj, jhryu\}@kaist.ac.kr}
    }
}

\maketitle

\begin{abstract}
High-definition (HD) maps, particularly those containing lane-level information regarded as ground truth, are crucial for vehicle localization research. Traditionally, constructing HD maps requires highly accurate sensor measurements collection from the target area, followed by manual annotation to assign semantic information. Consequently, HD maps are limited in terms of geographic coverage. To tackle this problem, in this paper, we propose SIO-Mapper, a novel lane-level HD map construction framework that constructs city-scale maps without physical site visits by utilizing satellite images and OpenStreetmap data. One of the key contributions of SIO-Mapper is its ability to extract lane information more accurately by introducing SIO-Net, a novel deep learning network that integrates features from satellite image and OpenStreetmap using both Transformer-based and convolution-based encoders. Furthermore, to overcome challenges in merging lanes over large areas, we introduce a novel lane integration methodology that combines cluster-based and graph-based approaches. This algorithm ensures the seamless aggregation of lane segments with high accuracy and coverage, even in complex road environments. We validated SIO-Mapper on the Naver Labs Open Dataset and NuScenes dataset, demonstrating better performance in various environments including Korea, the United States, and Singapore compared to the state-of-the-art lane-level HD map construction methods.

\end{abstract}


\section{Introduction}
\IEEEPARstart{A}{ccurate} vehicle localization is fundamental to most of autonomous driving algorithms such as path planning and control. However, these localization techniques mostly rely on \ac{HD} maps, which provide abundant semantic information including road topology, land topology, and traffic signs. Among the various information of \ac{HD} maps, in this paper, we focus on lane-level \ac{HD} maps, which offer precise lane information, which are particularly critical for vehicle localization methods.

% For most of autonomous driving algorithms such as path planning and control, determining an accurate position of the vehicle should be preceded. However, localization methods usually rely on \ac{HD} maps that provide abundant semantic information. \ac{HD} maps are composed of various information including point cloud, color image, and semantic information such as road topology, lane topology, and traffic signs. Among them, in this paper, we focus on lane-level \ac{HD} maps which is particularly popular for vehicle localization methods.

\begin{figure}[!t]
\centerline{\includegraphics[width=0.98\linewidth]{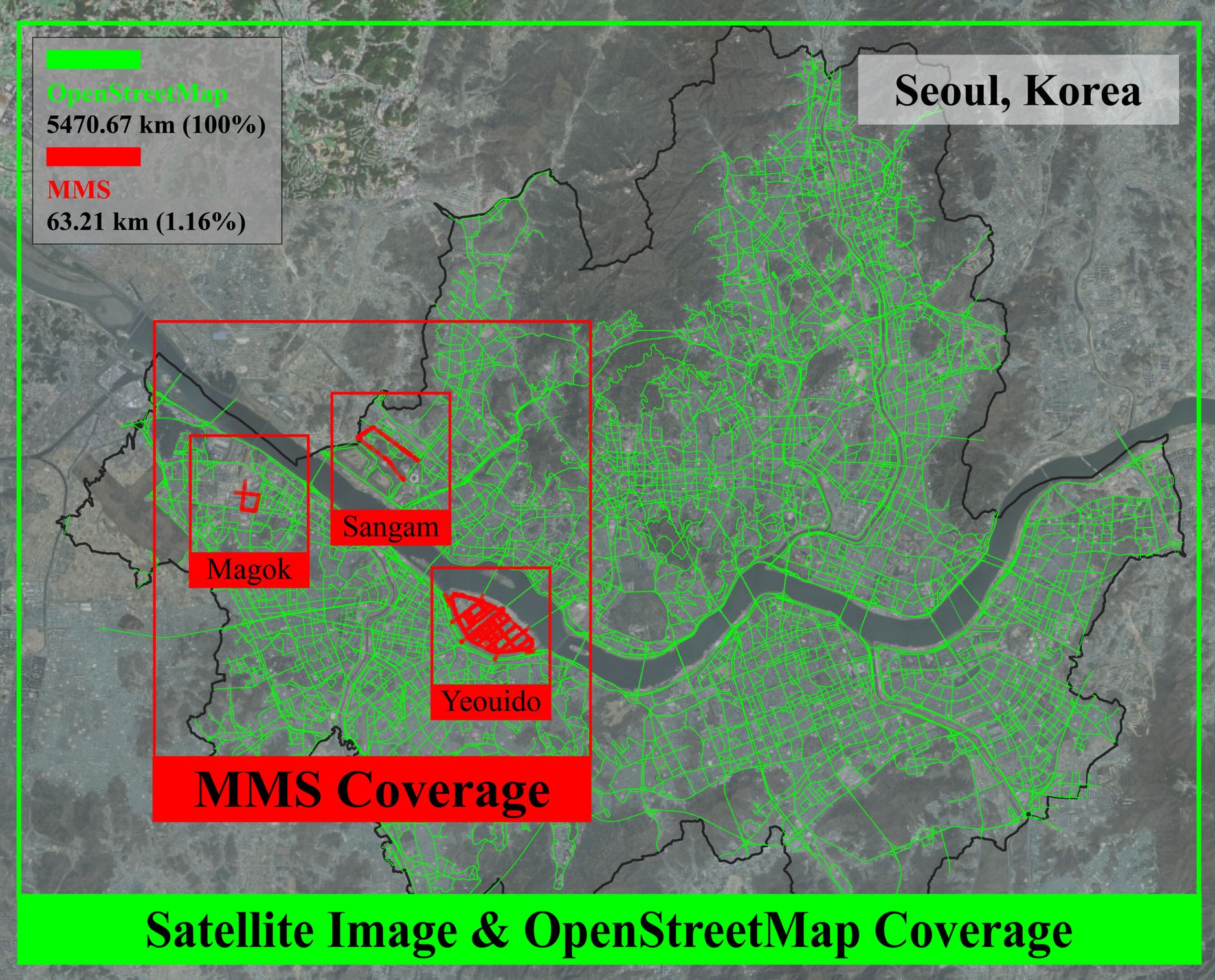}}
\caption{Conventional lane-level HD maps only cover very limited portion of the entire road. For example, in Seoul which contains 5470.67 km of roads, only 63.21 km which is 1.16\% of the road is covered as shown in this figure.}
% \caption{Conventional lane-level HD map is constructed based on sensor measurements that are acquired by mobile mapping systems. As a result, lane-level HD maps are only constructed in very limited area. For example, in Seoul which contains 5470.67 km of roads, only 63.21 km which is 1.16\% of the road is covered as shown in this figure. Therefore, to overcome this problem, this paper suggests a novel lane-level HD map construction framework that utilizes satellite image and OpenStreetMap road information which are publicly available online. Finally, we aim to construct lane-level HD maps everywhere without visiting the target area physically.}
% \vspace{-0.5cm}
\label{fig:intro}
\end{figure}

However, lane-level \ac{HD} map construction is currently limited to small geographic area due to the conventional methods that heavily rely on \ac{MMS} equipped with various sensors such as GPS, IMU, LiDAR, and cameras. \ac{MMS} acquires highly accurate sensor data but require physical visit to the target areas. Additionally, even if sophisticated mapping algorithms such as a state-of-the-art SLAM algorithm integrates sensor data into maps, significant human intervention is required to assign semantic information such as road and lane topology. Consequently, lane-level \ac{HD} map construction has been restricted to certain regions, and completed only by government agencies and large companies. For example, as shown in \figref{fig:intro}, Naver Labs Open Dataset dataset and NuScenes provide lane-level \ac{HD} map of Seoul and Boston seaport area but they only cover 63.21 kilometers and 21.91 kilometers which is about 1.16\% and 1.50\% of the entire road of each region as shown in \tabref{tab:coverage_ratio} \cite{naverlabs, nuscenes}.

% Unfortunately, lane-level \ac{HD} maps are only available in very limited area currently because conventional \ac{HD} map construction methods heavily rely on highly accurate sensor measurements acquired by visiting the target area of a \ac{MMS} equipped with various sensors such as GPS, IMU, LiDAR, and camera. Moreover, even if sensor measurements are integrated using mapping algorithms such as SLAM, human intervention is required to assign semantic information such as road topology and lane topology. As a result, only government agencies and large companies are possible to construct \ac{HD} maps for very limited area. For example, as shown in \figref{fig:intro}, Naver Labs Open Dataset dataset and NuScenes provide lane-level \ac{HD} map of Seoul and Boston seaport area but they only cover 63.21 kilometers and 21.91 kilometers which is about 1.16\% and 1.50\% of the entire road of each region as shown in \tabref{tab:coverage_ratio} \cite{naverlabs, nuscenes}.

\begin{table}[!t]
\caption{HD Map Coverage in Seoul and Boston}
\begin{center}
\resizebox{\linewidth}{!}{
\begin{tabular}{c|cc|c}
\midrule
Region & Total Road Length (km) & HD map Coverage (km) & Ratio \\
\midrule
Seoul  & 5470.67 & 63.21 & \textbf{1.16\%} \\
Boston & 1504.22 & 21.91 & \textbf{1.46\%} \\
\bottomrule
\end{tabular}
}
\label{tab:coverage_ratio}
\end{center}
% \vspace{-0.5cm}
\end{table}
\begin{figure*}[!h]
\centerline{\includegraphics[width=0.95\linewidth]{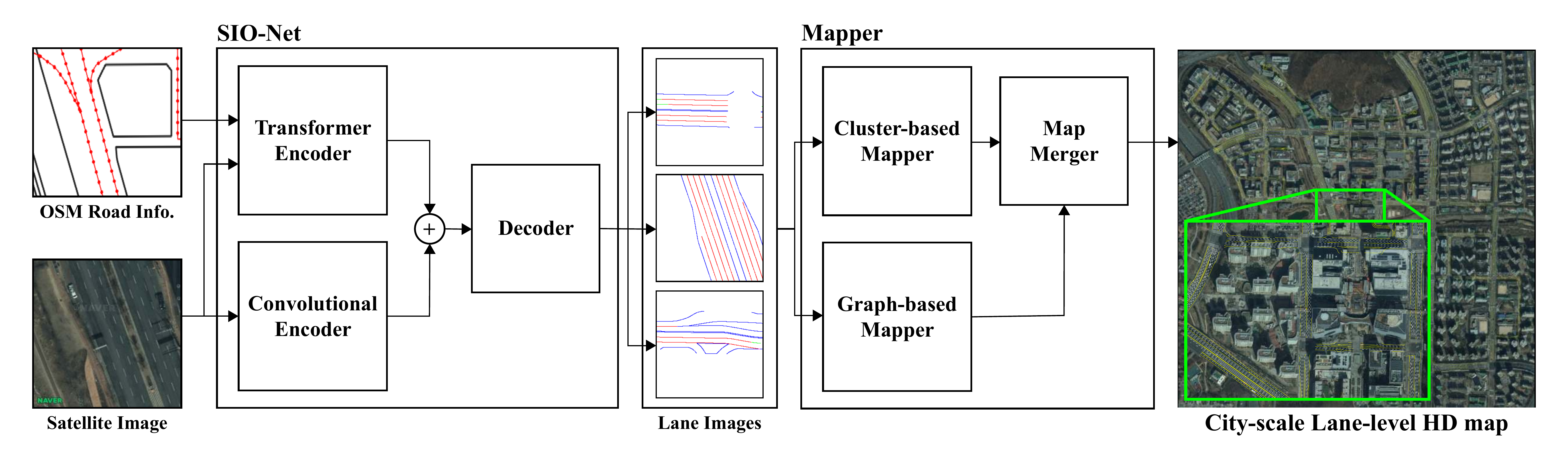}}
\caption{Overall pipeline of the proposed SIO-Mapper. First, SIO-Net generate lane images. Transformer encoder and convolutional encoder each extract features in satellite images. Then, decoder concatenate both features and generate lane images. Then, mapper integrates lane images using cluster-based mapper and graph-based mapper and merge them using map merger to construct lane-level HD map. Using SIO-Mapper, we can construct a lane-level HD map in city-scale without any physical visit to the target area.}
% \vspace{-0.5cm}
\label{fig:pipeline}
\end{figure*}

% Researchers have attempted to resolve the above two issues: automating the lane-level \ac{HD} construction process to eliminate human intervention [3-15] and constructing lane-level \ac{HD} maps without visiting the target location from public data sources instead of utilizing sensor measurements obtained from the \ac{MMS} [17-25]. Despite these efforts, researches still face a common limitation that they can only construct local \ac{HD} maps that are restricted to the sensor's scanning range.

Researchers have attempted to resolve two key challenges in lane-level HD map construction: automating the lane-level \ac{HD} construction process to eliminate human intervention and constructing lane-level \ac{HD} maps without visiting the target location from public data sources instead of utilizing sensor measurements obtained from the \ac{MMS}. The first approach [3-15] focuses on methods that can extract lane-level information from sensor data collected by \ac{MMS}. However, these methods often require costly, specialized equipment and are limited to the areas where sensors can be deployed, leading to restricted geographic coverage and high operational costs. On the other hand, the second approach [17-25] aims to construct lane-level HD maps using publicly available data, such as satellite images and OpenStreetMap (OSM), instead of on-site sensor measurements. While this method eliminates the need for physical site visits, existing techniques struggle to accurately stitch lane segments over large areas due to the lower resolution. Both approaches are still limited by the local scope of the maps they generate, as they rely on data with either narrow sensor range or incomplete coverage, making it difficult to construct seamless city-scale or global HD maps.

Even methods based on satellite imagery also suffer the same range limitation because high zoom levels are required to extract accurate lane information from the satellite imagery. These approaches struggle to integrate lane information over larger areas and only available to construct fragmented maps which lack the global context which are essential for city-scale autonomous driving. Moreover, most of conventional localization algorithms rely on comparing a query sensor measurement with reference maps, but existing \ac{HD} maps construction methods do not offer comprehensive information in city-scale coverage. Therefore, a method that can overcome these geographic limitations is required to enable large-scale autonomous driving in real situations.

% Prior researches attempted to overcome this problem by approaching in two different perspectives: an automated lane-level \ac{HD} construction method without human intervention [3-15] and construction of lane-level \ac{HD} map without visiting the target site by utilizing public data sources instead of sensor measurements [17-25]. However, these researches still share a common limitation that they only construct a local \ac{HD} maps that only covers within sensor range. Even methods using satellite images are only limited in the range of zoomed satellite image which is similar to the range of camera and LiDAR. Considering that most of conventional localization algorithms determine the position by comparing a query sensor measurement to reference map, a global \ac{HD} map construction method that covers city-scale regions is imperatively required.

In this paper, we propose SIO-Mapper, a novel lane-level \ac{HD} map construction framework designed to address these limitations. By utilizing publicly available satellite images and OpenStreetMap road data, SIO-Mapper constructs city-scale lane-level \ac{HD} maps without requiring sensor measurements from physical site visit. SIO-Mapper improves lane-level \ac{HD} map construction framework in two key modules as shown in \figref{fig:pipeline}. First, we propose SIO-Net, \textit{\textbf{S}}atellite \textit{\textbf{I}}mage and \textit{\textbf{O}}penStreetmap \textit{\textbf{Net}}work, a deep learning network that extracts lane information more effectively than existing networks by combining Transformer-based encoders and convolutional encoders. SIO-Net utilizes OpenStreetMap road data to guide lane placement and extract detailed lane information from satellite images which enables better lane extraction performance in both terms of accuracy and coverage. Second, we introduce a novel lane integration methodology that combines clustering-based and graph-based lane aggregation algorithms to merge lane segments across wider geographic areas. This approach addresses the limitations of merging lane segments in complex lane structures, ensuring the accurate lane-level \ac{HD} map construction in city-scale. As a result, SIO-Mapper not only expands the coverage but also enhance the precision and usability of the lane-level \ac{HD} maps, making it more effective for autonomous driving techniques in real situations. SIO-Mapper contributes in following three branches:

% Therefore, in this paper, we propose a novel lane-level \ac{HD} map construction framework that utilizes satellite images and OpenStreetMap road information which are publicly available for most of the roads without visiting the actual site. We divide and conquer the lane-level \ac{HD} map construction process into two module: extract lane from satellite image using the suggested neural network named SIO-Net which is an abbreviation for \textit{\textbf{S}}atellite \textit{\textbf{I}}mage and \textit{\textbf{O}}penStreetmap \textit{\textbf{Net}}work, and construct a global lane-level \ac{HD} map by stitching the extracted lane images. The first module overcomes the necessity of actual visits by utilizing only public data sources and the second module overcomes the range limitation by suggesting the novel lane merging algorithm that combines cluster-based and lane-based lane merging algorithms. As a result, our suggested framework is able to construct a global lane-level \ac{HD} maps without visiting the target mapping place and collecting sensor measurements. \figref{fig:pipeline} illustrates brief pipeline of suggested \ac{HD} lane mapping framework. Our paper contributes in following three branches:

\begin{itemize}

\item \textbf{Enhaned Lane Extraction using SIO-Net}\\ We introduce SIO-Net, a novel deep learning network that outperforms existing lane extraction networks. Unlike existing methods, which often struggle with limited detail and precision, SIO-Net utilizes OpenStreetMap road data to guide the lane extraction by providing road shape into the network. Furthermore, the combination of Transformer-based and convolutional encoders enables to find complex structural patterns and fine-grained lane details, respectively. These improvements reduce lane extraction errors and enhance accuracy which are more suitable for the real-world road network.

\item \textbf{City-scale Global HD Map Construction}\\ We also developed a city-scale lane-level \ac{HD} map construction framework that overcomes the geographical limitation of existing mapping methods which are mostly resrticted to the sensor's range. We achieve this by introducing a novel lane integration method that combines both clustering-based and graph-based methods. This approach selects only appropriate lanes for both methods, allowing for accurate lanes even in complex urban environments such as branching and merging lanes. The result offers significant improvements in both coverage and accuracy compared to existing methods. To the best of our knowledge, this is the first approach to construct a city-scale lane-level HD map without acquiring sensor measurements from physical site visit.

\item \textbf{Standardized Lane-level HD Maps Evaluation Metrics}\\ Finally, we contribute to the field by providing evaluation metrics specifically designed for lane-level HD maps. Current evaluation methods often focus on pixel accuracy, which does not capture the performance of lane continuity and connectivity that is crucial in real-world autonomous driving scenarios. Our metrics evaluate the structural quality of lane connection and map coverage, offering a standardized benchmark for future research in this domain.

\end{itemize}

\section{Related Work}
Lane-level \ac{HD} map construction has actively researched with approaches broadly falling into two categories: automating lane-level \ac{HD} map construction methods and lane-level \ac{HD} map construction without visiting the actual site by utilizing public data sources.

% In this section, we review prior researches on lane-level \ac{HD} map construction. We categorize and analyze them into two different categories, based on their relevance to the main challenges mentioned earlier: automating lane-level \ac{HD} map construction methods and lane-level \ac{HD} map construction without visiting the actual site by utilizing public data sources instead of sensor measurements.

\subsection{Automating Lane-level HD Map Construction Methods}
Researches to automate lane-level \ac{HD} map construction began with synthesizing onboard sensor measurements and extracting lanes from them. Early approaches mostly utilized cameras as their main sensor and also utilized passive sensors such as GPS and IMU additionally. For example, \citeauthor{guo2014automatic} integrated GPS and IMU data to extracted road segments based on the road network topology and accumulated \ac{BEV} images to generate synthetic orthographic images of the segment. From those synthetic images, they finally extracted lane graph and validates localization performance using them \cite{guo2014automatic, guo2016low}.

%%%%%% Previous Related Works %%%%%%

% As mentioned above, one of the main challenges is removing human intervention. Research on this problem has studied depending on the type of on-board sensor used. Early approaches mostly utilized cameras as their main sensor. \citeauthor{guo2014automatic} integrated GPS and IMU data to extracted road segments based on the road network topology and accumulated \ac{BEV} images to generate synthetic orthographic images of the segment. From those synthetic images, they finally extracted lane graph and validates localization performance using them \cite{guo2014automatic, guo2016low}.

With the advent of LiDAR sensors, which provide dense 3D information, the momentum of research shifted from cameras to LiDAR. \citeauthor{gwon2016generation} accumulated ground points from LiDAR measurements using GPS and INS then extracted road markings by simple intensity threshold. They completed the lane map by interpolating road markings using B-spline algorithm \cite{gwon2016generation}. \citeauthor{joshi2015generation} additionally utilized \ac{OSM} information as prior. They first extracted lane markings from LiDAR using intensity and aligned those lane markings into lines using RANSAC. Then, they performed particle filter along \ac{OSM} nodes. Finally, by optimizing number of lanes as same as \ac{OSM} information, they indexed the extracted lanes \cite{joshi2015generation}.

Then, research have advanced significantly with the development of deep learning networks. \citeauthor{liang2019convolutional} extracted road boundary from \ac{BEV} LiDAR, \ac{BEV} camera image, and elevation gradient derived from LiDAR using recurrent network named cSnake \cite{liang2019convolutional}. Similarly, \citeauthor{homayounfar2019dagmapper} introduced DAGMapper, which extracts lanes in directed acyclic graph format from LiDAR intensity image, using three recurrent networks \cite{homayounfar2019dagmapper}. \citeauthor{zurn2021lane} additionally utilized semantic information. They introduced LaneGraphNet, which generate directional lane graph from \ac{BEV} LiDAR, \ac{BEV} camera image, vehicles and semantics,  based on Graph-RCNN \cite{zurn2021lane, yang2018graph}. \citeauthor{zhou2021automatic} tried to complete lane graph by utlizing \ac{OSM} information. They extracted lane using DeepLabv3+ \cite{chen2018encoder} and particle filter, and then completed lane connection at intersections using \ac{OSM} road connecting topology \cite{zhou2021automatic}. \citeauthor{li2022hdmapnet} also suggested HDMapNet which construct local lane-level \ac{HD} maps within sensor range of a LiDAR and six surrounding images and \citeauthor{liu2023learning} detect 3D lanes from pseudo \ac{BEV} LiDAR image \cite{li2022hdmapnet, liu2023learning}. Finally, \citeauthor{ort2022maplite} introduced a remarkable HD lane mapping framework using \ac{OSM} centerline information as prior. They extracted lanes using also DeepLabv3 and updated \ac{OSM} centerline prior to complete lane-level \ac{HD} map, which are considered as state-of-the-art method \cite{ort2022maplite}.

% However, despite the advancement of various methods using variety of sensors, there is still an inherent problem that lane-level \ac{HD} maps can only be constructed by physically visiting the target mapping area.

\subsection{Lane-level HD Map Construction using Public Data Sources}
% Therefore, to avoid using on-board sensors that force visits and thus limit the coverage of lane-level \ac{HD} maps, researchers have become utilizing satellite images which are collectable without visiting the actual places.
To overcome the need for the physical visit of a \ac{MMS}, reseachers started to use satellite images which are publicly available and offer global coverage. \citeauthor{mattyus2017deeproadmapper} segmented road and non-road area using ResNet \cite{he2016deep} and generated road graph using A* algorithm \cite{mattyus2017deeproadmapper}. \citeauthor{azimi2018aerial} suggested Aerial LaneNet, FCNN-based neural network enhanced by \ac{DWTs}. They outperformed existing algorithms in terms of pixel accuracy and \ac{IoU} \cite{azimi2018aerial}. However, Aerial LaneNet returns an output as rasterized image. However, rasterized lane images are difficult to construct maps in larger regions, as they are pixel-based information which cannot efficiently represent the continuous and interconnected road networks. Therefore, vectorized lane maps are required to maintain lane continuity and connectivity across wide regions.

% which is almost impossible to integrate in global-scale and does not have semantic information of lanes such as connectivity. Therefore, to construct a global lane-level \ac{HD} maps, we need vectorized format.

From that perspective, \citeauthor{he2022lane} extracted lane-level street map using U-Net \cite{ronneberger2015u} with ResNet backbone. They divided and conquered this problem into non-intersection and intersection area. They first segmented lanes at non-intersection area and then connected them at intersections \cite{he2022lane}. \citeauthor{xu2021cp} has tried to solve this problem across a series of papers. They first detected road curbs which is similar with road boundary using two approaches: suggesting connectivity-preserving loss for U-Net and adopting \ac{FPN} \cite{xu2021cp, xu2021icurb, lin2017feature}. Finally, they constructed city-scale road boundary by adding \ac{AfANet} after \ac{FPN} to connect road boundaries by considering them as graph \cite{xu2022csboundary}. Also, they suggested RNGDet, which constructs road graph using Transformer \cite{xu2022rngdet, vaswani2017attention}.

\section{Methodology}

\begin{figure*}[!t]
\centerline{\includegraphics[width=0.95\linewidth]{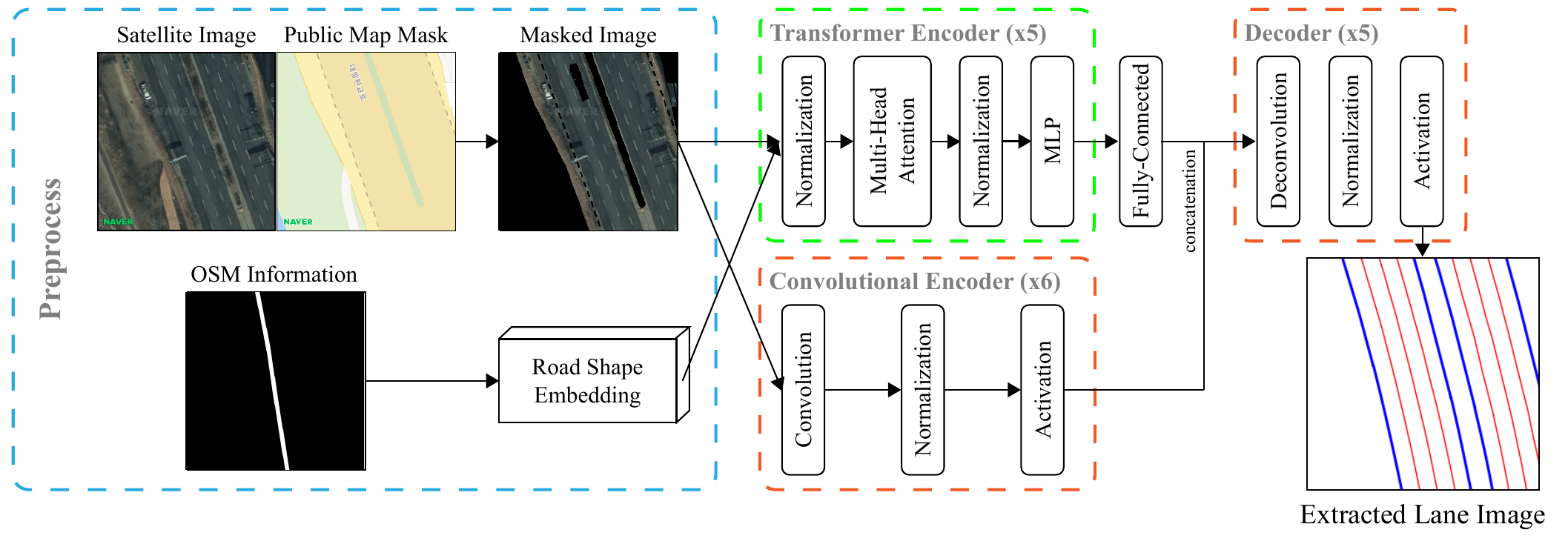}}
\caption{This figure illustrates the pipeline of the suggested network, SIO-Net. In preprocess step, we mask the satellite image using public map mask and calculate road shape embedding by approximating the shape of road into quadratic function. Then, we extract image features using two different encoders, Transformer encoder and convolutional encoder, and concatenate both features. Finally, the decoder generate a lane image by decoding the concatenated feature. Lane image is consists of three labels which are respectively color coded in red, green, and blue: white broken lanes, white lanes, and yellow lanes.}
\vspace{-0.5cm}
\label{fig:network}
\end{figure*}

To address these limitations, this section describes the detailed structure of proposed SIO-Mapper framework, which enables the city-scale lane-level \ac{HD} map construction using publicly available satellite imagery and OpenStreetMap road information, without requiring physical visits to the target areas. SIO-Mapper framework consists of two main modules: SIO-Net for the lane extraction and the mapper for lane integration. The first module, SIO-Net, processes satellite images to extract lane segments using a combination of Transformer-based and convolution-based encoders. Transformer encoder leverages OpenStreetMap road information, embedding road shapes to guide the lane extraction process. By combining features from both encoders, SIO-Net generates lane images that contains both the structural and visual properties of the road. The second module, mapper, integrates these extracted lane images into a lane-level \ac{HD} map. This is achieved through a novel algorithm that combines advantages of clustering-based and graph-based stitching method. While clustering is effective in lane discrimination, graph-based method is effective for lane connectivity especially in complex road layouts. Unlike existing approaches which struggle with lane connectivity, the suggested method allows for the seamless merging of lane images by selecting lanes that are appropriate for both methods' criteria. By using Hungarian algorithm to optimize the lane stitching process, SIO-Mapper ensures seamless and accurate integration of lane images into cohesive, lane-level \ac{HD} map that covers city-scale large geographic regions.

% This section describes the suggested lane-level \ac{HD} map construction framework, SIO-Mapper, in detail. As shown in \figref{fig:pipeline}, SIO-Mapper enables a lane-level \ac{HD} map construction without visit by utilizing public data sources only: satellite images and OpenStreetMap road information. SIO-Mapper is consists of two modules. The first module, SIO-Net, extracts first features from satellite image using Transformer encoder and convolutional encoder. For Transformer encoder side, OpenStreetMap road information assists the encoder by providing the road shape information as a feature embedding. Finally, by decoding both features, SIO-Net generates lane image from satellite image assisted by OpenStreetMap road information. The second module, mapper, integrates lane images and constructs a lane-level \ac{HD} map using a novel method that combines both advantages of clustering and graph-based stitching method. Each method have limitations in particular situation such as branching lanes. However, the suggested method overcomes those situations by finding the best fitting lanes which satisfies both criteria of clustering and graph-based stitching method using Hungarian algorithm.

% 여기가 중요한데... 기존 어떤 문제가 있었는데.. 두가지 정보를 상호보완적으로 사용해서 어떻게 해결할 수 있었다라는 식으로 좀더 상세하게 설명할 필요가 있음. 너무 불친절하게 작성되어 있네!!
\subsection{Problem Formulation}
This paper focuses to construct lane-level \ac{HD} maps without need for sensor-based site visits. Instead, we rely solely on publicly available data: satellite images and OpenStreetMap road information. These two data sources are complementary because OpenStreetMap provides global road network topology for wide areas, offering a structural skeleton, while satellite images offers detailed visual information for accurate lane extraction. The challenge lies in seamlessly combining these data to generate a coherent city-scale lane-level \ac{HD} map. To achieve this, our framework first extracts lane images using SIO-Net, which is guided by the OpenStreetMap road shape information. The extracted lane images are then integrated using a novel mapper that ensures connectivity and accuracy by combining clustering-based and graph-based methods, resulting in a comprehensive city-scale lane-level \ac{HD} map.

% This paper focuses on the construction of the lane-level \ac{HD} maps without visit by utilizing only satellite images and OpenStreetMap road information which are publicly available. The selection of satellite these two data sources is predicated on their ability to function complementarily in the lane-level \ac{HD} map construction process. OpenStreetMap road information serves as the foundational skeleton by offering road topology for expansive areas, whereas satellite images augment this framework by providing detailed lane information in the vicinity of road locations. Therefore, we first download satellite images following the roads of OpenStreetMap and extract lanes using the suggested network named SIO-Net. Then, finally, we aggregate extracted lanes and construct a lane-level \ac{HD} map in city-scale using the suggested mapper which combines cluster-based and graph-based method.

If represented as a formula, denoting satellite images $S = \{s_1, s_2, \cdots, s_n\}$ with their center at interpolated road locations from OpenStreetMap as $X = \{x_1, x_2, \cdots, x_n\}$, approximated road direction coefficients as $C = \{c_1, c_2, \cdots, c_n\}$ and SIO-Net as $\phi(\cdot) = D(E_t(\cdot), E_c(\cdot))$, where $E_t(\cdot, \cdot)$, $E_c(\cdot)$, and $D(\cdot, \cdot)$ denote Transformer encoder, convolutional encoder, and decoder respectively, lane images $L = \{l_1, l_2, \cdots, l_n\}$ are generated as $L_i = \phi(s_i, c_i)$. Then, the mapper $M(M_c(\cdot), M_g(\cdot))$, where $M_c(\cdot)$ and $M_c(\cdot)$ denote clustering-based and graph-based mapper respectively, constructs a lane-level \ac{HD} map $G$.

\subsection{Lane Extraction using SIO-Net}
To construct a reliable lane-level \ac{HD} map, the lane extraction process should not only be highly accurate at the pixel level but should also maintain lane continuity and connectivity across the entire map. Therefore, we introduce SIO-Net, a novel deep learning network that combines both advantages of Transformer-based and convolutional neural networks as shown in \figref{fig:network}. Transformer encoder focuses on connectivity of the road by embedding the road shapes provided by OpenStreetMap. Convolutional encoder, on the other hand, focuses on lane discrimination. By merging the output features of these two encoders, SIO-Net extracts lane images that are accurate in both terms of image accuracy and structural accuracy, addressing the critical need for lane continuity over city-scale areas.

% Considering that the purpose of the lane extraction is to construct a lane-level \ac{HD} map, the lane extraction result should be not only accurate in terms of image accuracy such as pixel accuracy and \ac{mIoU} but also represent the connectivity of each pixels which included in same lane. Therefore, we suggest a novel deep learning network names SIO-Net, which combines both advantages of Transformer and convolutional neural networks to guarantee both image accuracy and the connectivity \cite{vaswani2017attention}. Overall pipeline of the suggested network, SIO-Net, is shown in \figref{fig:network}.

\subsubsection{Preprocess}
In the preprocessing stage, we refine the satellite images to optimize the lane extraction process. To ensure the network focuses on relevant road areas, we filter out satellite images that do not contain any roads or lanes. Using OpenStreetMap data, we identify road locations and interpolate them at 1-meter intervals. Satellite images are then downloaded at these road points, with each image covering a 64m x 64m area at a resolution of 256 x 256 x 3 pixels. Also, because non-road area in satellite image also can interrupt the training, we mask the satellite image using public map downloaded at the same location.

Additionally, we preprocess the OpenStreetMap (OSM) road data, which is used as input to the Transformer encoder. To capture the road topology effectively, we approximate the shape of the roads using a quadratic function, which generates road shape coefficients. These coefficients are used as a road shape embedding in the Transformer encoder, providing structural guidance during the lane extraction process. This embedding acts as a reference to ensure that the extracted lanes align accurately with the underlying road topology, improving the consistency and accuracy of lane extraction across various environments.

% First, we preprocess satellite images to enhance the efficiency of network training stage. Since images which do not contain any lanes might inhibit the training, we only select satellite images at road area using OpenStreetMap road information. We interpolated OpenStreetMap road information with 1\,m interval and download satellite images $S$ at each road points $X$. Each satellite image $s_i$ have pixel size of $(256,256,3)$ which covers $(64\,m, 64\,m)$ in actual scale. Also, because non-road area in satellite image also can interrupt the training, we mask the satellite image using public map downloaded at the same location.

% Then, we also preprocess OpenStreetMap road information which used as input to the Transformer encoder $E_t$. To represent the road topology and use this information during the network training, we approximate the shape of roads at the certain position in a quadratic function and use coefficients $C$ as road shape embedding of Transformer encoder. Road shape embedding serves a role of a guide to ensure consistent lane extraction by guiding the generated lanes to follow the shape of the road.

\subsubsection{Transformer Encoder}
Transformer encoder $E_t$ extract features from satellite images based on Transformer and OpenStreetMap road information is used as road shape embedding. Transformer was originally designed as \ac{LLM} which is specialized in calculating self-attention. Inspired by Transformer, we designed an encoder to calculate attention between image patches, so that patches that share same lane have a higher attention score. By using $4\times4$ image patch, $(256,256,3)$ image is transformed in $(4096,784)$ shaped array and by adding three road topology coefficients $C$, the input shape becomes $(4099,784)$.

Transformer encoder $E_t$ is consist of five Transformer blocks. Because each Transformer block does not changes the shape of the input, an output feature $f_t$ of the Transformer encoder $E_t$ also has a shape of $(4099,784)$.

\subsubsection{Convolutional Encoder}
Convolutional encoder $E_c$ focuses on generating the accurate lane image $L$. In this module, we leveraged a conventional encoder structure consisting of convolution, normalization and activation functions.

Convolutional encoder $E_c$ is consist of six encoder blocks with 16, 16, 32, 64, 128, and 256 feature dimensions and each block halves image size in both width and height. As a result, convolutional encoder returns an output feature $f_c$ with a shape of $(256,4,4)$.

Because two encoders return features in different shape, we should match their shape before we feed them into the decoder. Using four convolutional layers that halve image size, a flatten layer, and a linear layer, we convert the output of Transformer encoder $f_t$ to have the same shape, $(256,4,4)$, as output of the convolutional encoder $f_c$.

\subsubsection{Decoder}
Finally, decoder $D$ generates a lane image $L$ from two features $f_t$ and $f_c$ from each encoder. The decoder is consists of five decoder blocks. The first decoder block concatenates both features. From the second blocks, each block concatenates the output of the previous decoder block and the output of convolutional encoder block at the same level. Then, the concatenated tensor passes through a deconvolution layer that doubles image size, normalization layer, and activation layer. Each of them have 128, 64, 32, 16, and 3 feature dimensions which is the reverse of feature dimensions of convolutional encoder $E_c$.

\subsection{Mapper}
To complete the lane-level \ac{HD} map $G$, we aggregated lane images $L$ generated by SIO-Net. The most intuitive method to aggregate lane images is to convert lane images into lane points on the global coordinate system and cluster them. Clustering algorithms are powerful and efficient method to group lane points, however they have a crucial disadvantage that a point can only belong to only one cluster. As a result, clustering algorithm cannot perfectly discriminate lanes in specific conditions such as merging and diverging lanes. On the other hand, a graph-based methods are also can be adopted by considering each lane image as a graph and merging them. Graph-based methods can overcome the aforementioned problem, however they heavily rely on hyperparameters and also need some heuristics. Therefore, we introduce a novel mapper that takes only advantages of both clustering-based and graph-based method.
% 그림을 새로 그리기 (mapper의 pipeline) + fig3도 수정해야함

% \input{figures/fig4_clustering.tex}

% \subsubsection{Preprocessing}
% Before the \ac{HD} lane map construction, we need to preprocess \ac{OSM} and satellite images. Because road takes a very small portion of the satellite images, we indicate the road area using \ac{OSM} road information. However, 'lines' layer in \ac{OSM} contains variety of information, including not only roads, but also buildings and rivers. We extract only road information labeled as 'primary', 'secondary', 'residential', 'tertiary', and 'trunk'. Furthermore, they consists of minimum points that can represent the shape of roads, so we interpolate them to create road points. Then, for each road points, we crop the satellite image to $256\times256$, which is the input size of the suggested network. Finally, we generate lane images by using prepared satellite images and aforementioned network. As a result, for each road, we generate lane images $L = \{L_1, L_2, \cdots, L_n\}$.

% problem definition 먼저 하기, 거기서 나오는 notation 바탕으로 쓰기 (SIO-Net 부분도 수정)

\subsubsection{Clustering-based Mapper}
First, clustering-based mapper $M_c$ starts with converting each lane image $l_i$ into lane points $P = \{(x_1, y_1), (x_2, y_2), \cdots, (x_n, y_n)\}$ using \eqref{eq:image2point}.

\begin{equation}
\begin{split}
% \tensor[^G]{x}{} &= x_R + p \cdot (\tensor[^P]{x}{} - \frac{w}{2}) \cdot \cos{(\theta_R)} \\
% \tensor[^G]{y}{} &= y_R + p \cdot (\tensor[^P]{y}{} - \frac{h}{2}) \cdot \cos{(\theta_R)},
x_j &= x_i + p \cdot (\tensor[^P]{x}{} - \frac{w}{2}) \cdot \cos{(\theta)} \\
y_j &= y_i + p \cdot (\tensor[^P]{y}{} - \frac{h}{2}) \cdot \cos{(\theta)},
\end{split}
\label{eq:image2point}
\end{equation}

where $(x_j, y_j)$ denotes a global coordinate of each lane point, $(x_i, y_i)$ denotes a global coordinate of image center which is road location from OpenStreetMap, $p$ denotes pixel size in meter scale which is 0.25\,m in this paper, $(\tensor[^P]{x}{}, \tensor[^P]{y}{})$ denotes a pixel coordinate of a lane point, $(w,h)$ denotes image size, and $\theta$ denotes angle between image center and target pixel. Using this coordinate conversion, we generate lane points $P_i$ that is according to each lane image $l_i$.

Then, we cluster lane points using DBSCAN \cite{ester1996density} where each cluster represents each lane. In this module, we aggregate every lane points which belongs in same \ac{OSM} road and cluster them at once.

\begin{gather}
\bar{P} = \cup_{i=1}^{n}{P_i} \\
DB(\bar{P}, 6, 10) = C = \{c_1, c_2, \cdots, c_n\},
\label{eq:clustering}
\end{gather}

where $DB(\cdot,\cdot,\cdot)$ denotes DBSCAN algorithm and each element denotes target points, minimum number of points, and maximum distance value used for clustering. Each cluster $c_i$ represent lanes which are aggregated using clustering-based mapper. As a result, lane images $L$ generate lanes $C = \{c_1, c_2, \cdots, c_n\}$ in cluster form.

\vspace{0.2cm}

\subsubsection{Graph-based Mapper}
On the other hand, graph-based mapper first cluster lane points $P_i$ in each lane image $l_i$ using $DB(P_i, 4, 5)$. Then, we convert each lane image $l_i$ into a graph $g_i$ by assigning each lane cluster as a edge, and farthest point pairs of each cluster as start and end vertices of the lane. Then, we merge edges if their vertices are adjacent within 1\,m. As a result, we get a graph $g_i = \{e_1, e_2, \cdots, e_n\}$ which is composed of merged graph edges. Each edge $e_j$ represent lanes which are aggregated using graph-based mapper. As a result, lane images $L$ generate lanes $E = \{g_1, g_2, \cdots, g_n\}$ in vectorized form.

% 수식으로 표현 필요 (뒤에 나오는 E_j와 연결)

\subsubsection{Map Merger}
Finally, we merge lane clusters $C$ and lane graphs $E$ to construct a lane-level \ac{HD} map. Using the equation \eqref{eq:hungarian}, we find the cluster-edge pairs that maximizes the sum of the sharing points in each pair. This procedure enables to construct a global lane-level \ac{HD} map that preserves both lane discrimination ability of clustering-based mapper and lane connectivity of graph-based mapper. By denoting the weight matrix $\omega$, where $\omega(i,j)$ represents number of sharing lane points between lane cluster $c_i$ and lane graph edge $g_j$, we maximize total number of sharing lane points using Hungarian algorithm as shown in following equations. Finding the permutation matrix $P$ that maximizes the sum of the trace of $P\omega$ has the exactly same physical meaning with finding the best pair that maximizes the sum of the sharing points.

% \begin{equation}
\begin{gather}
\omega(i,j) = |C_i \cap E_j| \\
P = \argmax_P{\sum{Trace(P\omega)}},
\end{gather}
\label{eq:hungarian}
% \end{equation}
% PPT 참고해서 그림 완성하기
% clustering -> 무지개 색으로 확실히 구분되도록 표현하기 
% MATLAB으로 그려야하는 그림만 남은 상태

where permutation matrix $P$ designate the cluster-edge pair $\Tilde{P} = \{(\Tilde{c_1}, \Tilde{e_1}), (\Tilde{c_2}, \Tilde{e_2}), \cdots, (\Tilde{c_n}, \Tilde{e_n})\}$ that maximizes sharing lane points. Finally, we can construct a lane-level \ac{HD} map $G = \{\Tilde{c_1}, \Tilde{c_2}, \cdots, \Tilde{c_n}\} = \{\Tilde{g_1}, \Tilde{g_2}, \cdots, \Tilde{g_n}\}$ that maximizes both advantages of clustering-based mapper and graph-based mapper. As a result, a global lane-level \ac{HD} maps which have city-scale lane information can be constructed various environments using this framework as shown in \ref{chap:results}.
\section{Experiment}

\subsection{Benchmark Dataset}
We validated our lane-level \ac{HD} map construction framework on eight sequences from two datasets which provides \ac{HD} lane information: Naver Labs Open Dataset \cite{naverlabs} and NuScenes Dataset \cite{nuscenes}. We directly utilized provided vectorized lane information as ground truth lane-level \ac{HD} map and converted them into lane images and utilized as deep learning labels. Detailed information of datasets are described in \tabref{tab:dataset}.

\begin{table}[!t]
\caption{Information of Datasets used in This Experiment}
\begin{center}
\resizebox{\linewidth}{!}{
\begin{tabular}{ccccccccc}
\midrule
\multirow{2}[3]{*}{\shortstack{Dataset\\ (Sequence)}} & \multicolumn{4}{c}{NAVER LABS Open Dataset} & \multicolumn{4}{c}{NuScenes} \\ \cmidrule(rl){2-5} \cmidrule(rl){6-9}
 & Pangyo & Sangam & Yeouido & Magok      & Boston & One North & Holland & Queens \\
\midrule
\# Road        & 111    & 101    & 189    & 110    & 164    & 720    & 263    & 341 \\
\# Road Points & 18,299 & 11,571 & 24,718 & 15,117 & 22,201 & 20,941 & 9,074 & 12,083 \\
Length (km)    & 20.32  & 15.24  & 31.81  & 16.16  & 21.91  & 25.56  & 11.98  & 16.16 \\
\bottomrule
\end{tabular}
}
\label{tab:dataset}
\end{center}
% \vspace{-0.5cm}
\end{table}

\subsection{Comparison Method}
To evaluate our suggested network, SIO-Net, we compared our result with HDMapNet\cite{li2022hdmapnet}, which is considered as state-of-the-art \ac{HD} lane extraction algorithm, utilizing on-board sensor measurements: six surrounding images and one LiDAR scan. By comparing with HDMapNet, we validated our suggested network shows comparable result with existing methods with only public data sources. However, because Naver Labs Open Dataset provides only four surrounding images, we revised the HDMapNet to use four images and trained the network again. Also, result of Yeouido sequence is not included because Naver Labs Open Dataset does not provide sensor measurements for the Yeouido sequence. Please note that SIO-Net distinguishes between three types of lanes, while HDMapNet extracts lanes indifferently.

On the other hand, to evaluate our suggested lane-level \ac{HD} map construction framework, SIO-Mapper, we compared our resulting lane-level \ac{HD} map with lane information provided by Naver Labs Open Dataset and NuScenes Dataset using after-mentioned lane-level \ac{HD} map evaluation metrics we suggest.

\subsection{Training Details}
We sampled 69,705 satellite images that correspond to Naver Labs Open Dataset lane information. We utilized 55,764 images for training and 13,941 images for validation. We trained our suggested network on four Titan V GPUs over 10 epochs. On the other hand, HDMapNet was trained over 30 epochs with same condition for all other conditions.

Also, we adopted a loss function $L = L_R + L_S$, where $L_R$ denotes reconstruction loss and $L_S$ denotes semantic loss. For $L_R$, we used \ac{BCE} loss and for $L_S$, we introduced a novel semantic loss which represents the difference between the number of lanes in the label and number of clusters in generated lane image. This semantic loss function accelerates the training sequence by forcing each lane to become less blurred and clearer to match the number of lanes.

\subsection{Lane-level HD map Evaluation Metrics}
Evaluation lane-level \ac{HD} maps, especially on a global scale, requires more than conventional metrics such as pixel accuracy and \ac{mIoU}, which are mostly used for image evaluations. These traditional metrics fall short when evaluating maps in vector format, where lane continuity and connectivity are crucial. To provide a comprehensive evaluation, we propose three new quantitative metrics specifically designed to assess the quality of vectorized lane-level \ac{HD} maps.

% Researches on a global-scale lane-level \ac{HD} map construction methods are not popular and even they are using non-unified evaluation metric to validate the suggested methods. Conventional lane-level \ac{HD} map construction methods usually extracted lane information into image form that cover within sensor range. Therefore, they mostly utilized pixel accuracy and \ac{mIoU}, which are known as metric of image similarity, for their evaluation. However, accuracy in terms of image similarity is not proper to evaluate a lane-level \ac{HD} map, which are mostly constructed in vectorized format. Therefore, in this paper, we propose three quantitative criteria to evaluate vectorized lane-level \ac{HD} map.

\vspace{0.5cm}

\textbf{Map Coverage (}\boldsymbol{$C_{m}$}\textbf{):} $C_{m}$ indicates how much of the target mapping area is covered by the constructed lane-level \ac{HD} map in a percentage. To match constructed lane-level \ac{HD} map, $\Tilde{M}$, and ground truth lane-level \ac{HD} map, $\bar{G}$, we utilized Hungarian method again.

This time, $\omega(i,j)$ denotes the distance between start and end vertices of $\Tilde{\bar{G}}_i$ and $\bar{G}_j$, where $\Tilde{\bar{G}}_i$ and $\bar{G}_j$ represents edge $i$ of $\Tilde{\bar{G}}$ and edge $j$ of $\bar{G}$, respectively. By denoting $P^{\prime}=P\omega$ and number of edges in $\bar{G}$ as $|\bar{G}|$, we calculate a coverage of constructed lane-level \ac{HD} map as shown in \eqref{eq:map_cov}. 

\begin{equation}
C_m = \frac{|Trace(P^{\prime})>0|}{|M|} \times 100 (\%)
\label{eq:map_cov}
\end{equation}

On the other hand, we evaluate how accurate the map is using two criteria: map accuracy $A_{m}(\theta)$ and mean vertex distance $D_{v}$.

\textbf{Map Accuracy (}\boldsymbol{$A_{m}(\theta)$}\textbf{):} First, accuracy is measured by how many lanes are constructed within the threshold compared to the ground truth. As shown in \eqref{eq:map_acc}, we count number of edge pair previously bounded by Hungarian method and also calculate $A_m(\theta)$ in a percentage.

\begin{equation}
A_m(\theta) = \frac{P^{\prime}(\theta)}{|M|} \times 100 (\%), \quad P^{\prime}(\theta) = |Trace(P^{\prime})<\theta|
\label{eq:map_acc}
\end{equation}

\textbf{Mean Vertex Distance (}\boldsymbol{$D_{v}$}\textbf{):} Finally, mean vertex distance, $D_v$, describes the actual distance, in meter scale, between the constructed lanes and ground truth lanes.

\begin{equation}
D_v = \frac{\sum(Trace(P^{\prime}))}{|Trace(P^{\prime})>0|} (m)
\label{eq:node_dist}
\end{equation}
%%%%%%%%%%%%%%% TODO %%%%%%%%%%%%%%%
% Result 글 전체적으로 수정
% Mapper pipeline 그림 그리기 (한 단 사이즈)
% 결과 그림 detail, overview 모두 추가하기 (발표 slide p.46)

\section{Results}
\label{chap:results}
In this section, we present the result of the proposed method. First, we evaluate the lane extraction performance of our suggested network, SIO-Net, against to HDMapNet, which is considered as a state-of-the-art algorithm using on-board sensor measurements. Second, we evaluate the constructed lane-level \ac{HD} map using SIO-Mapper using the suggested lane-level \ac{HD} map evaluation metrics.

\subsection{SIO-Net Evaluation}
\begin{figure}[!t]
\centerline{\includegraphics[width=0.95\linewidth]{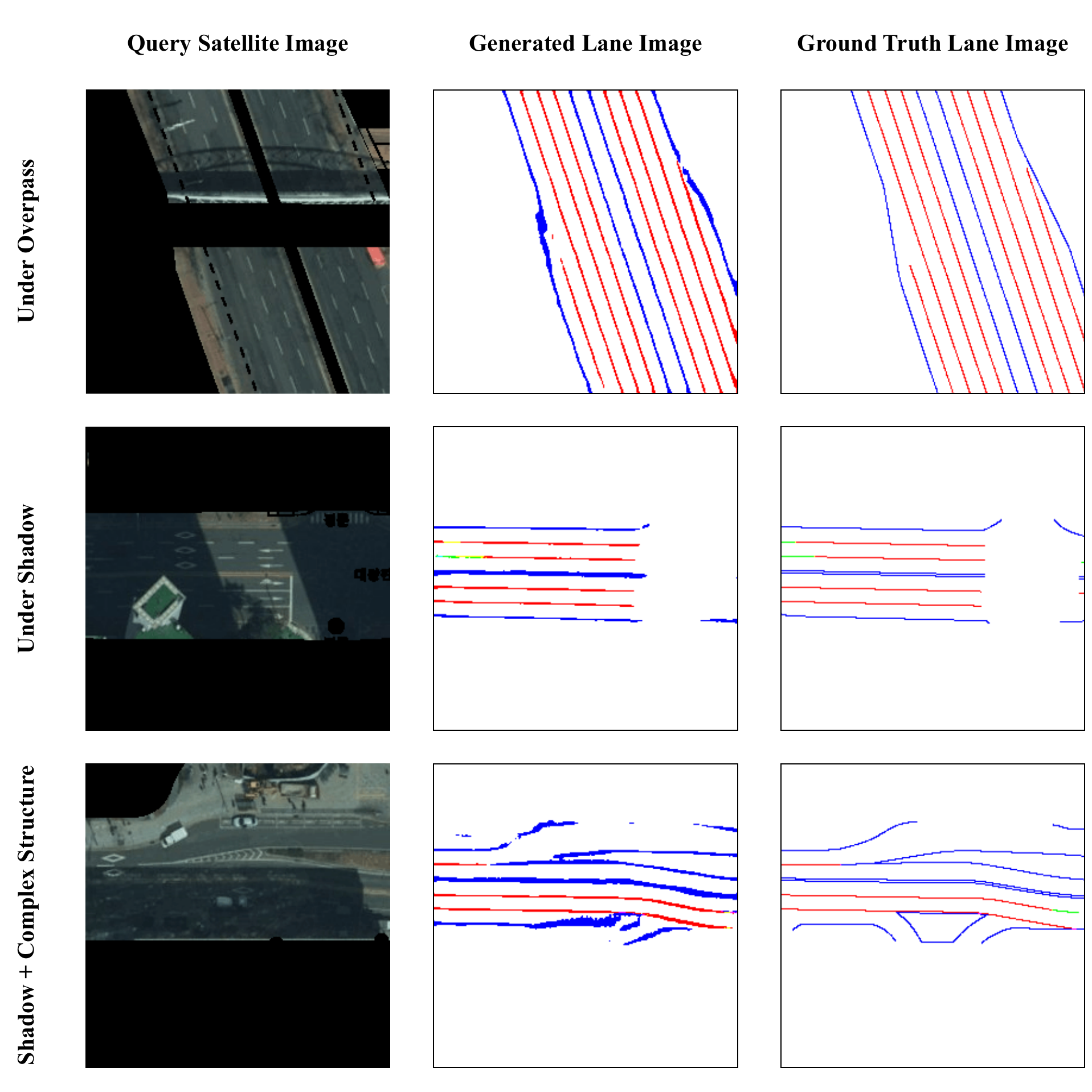}}
\caption{Result at three challenging cases: under overpass, under shadow, and complex lane structure. Even a query image is partially blocked by diverse obstacle, OST still can generate lane image well. The last case, complex lane structure, does not perform perfectly, but they still perform properly except road boundary, where the vehicle actually drives.}
\vspace{-0.5cm}
\label{fig:net_result}
\end{figure}

First, we evaluated SIO-Net on four sequences of Naver Labs Open Dataset. \figref{fig:net_result} illustrates three example cases in challenging environment: lanes under overpass, lanes under shadow, and a complex lane structure. As \figref{fig:net_result} shows, SIO-Net generates lane images with high accuracy even under challenging conditions. \tabref{tab:pixel_acc} details performance of SIO-Net compared to HDMapNet in terms of a pixel accuracy. Because HDMapNet does not discriminate types of lanes, we only calculated total pixel accuracy while we calculated pixel accuracy for each lane type for SIO-Net. Our proposed network performed better than HDMapNet utilizing only public data sources, achieving over 99\% accuracy for each of the four sequences.

\tabref{tab:miou} describes $mIoU$ of both SIO-Net and HDMapNet. As \tabref{tab:miou} shows, SIO-Net achieved significantly lower $mIoU$ compared to both HDMapNet and pixel accuracy of SIO-Net. The reason for the low $mIoU$ is that lanes only take a very small portion of the lane image, approximately 1\%, as shown in the \tabref{tab:miou}. While lanes in ground truth lane image perfectly have width of 1 pixel, lane images generated using deep learning network have thicker lanes. Therefore, union pixels becomes considerably larger than the intersection pixels and this phenomenon makes $mIoU$ smaller because $mIoU$ is calculated as intersection over union. However, in cluster-based and graph-based mapper, thicker lanes becomes a cluster and a graph edge so it does not effect to the resulting lane-level \ac{HD} map.

\begin{table}[!t]
\caption{Pixel Accuracy of \textit{SIO-Net}}
\begin{center}
\resizebox{\linewidth}{!}{
\begin{tabular}{|c|c|cccc|c|}
\hline
\multirow{2}{*}{\textbf{Network}}          & \multirow{2}{*}{\textbf{Lane Type}} & \multirow{2}{*}{Pangyo} & \multirow{2}{*}{Sangam} & \multirow{2}{*}{Yeouido} & \multirow{2}{*}{Magok} & \multirow{2}{*}{\textbf{Average}} \\
&&&&&&\\
\hline
\multirow{2}{*}{\textbf{HDMapNet}}         & \multirow{2}{*}{\textbf{Total}}         & \multirow{2}{*}{87.61\%} & \multirow{2}{*}{91.85\%} & \multirow{2}{*}{-} & \multirow{2}{*}{44.68\%} & \multirow{2}{*}{74.71\%} \\
&&&&&&\\
\hline
\multirow{4}{*}{\textbf{\textit{SIO-Net}}} & White, Broken  & 99.23\% & 99.32\% & 99.38\% & 99.58\% & 99.38\% \\
                                           & White          & 99.85\% & 99.88\% & 99.89\% & 99.94\% & 99.89\% \\
                                           & Yellow         & 98.68\% & 98.60\% & 98.34\% & 98.83\% & 98.61\%\\
                                           & \textbf{Total} & \textbf{99.26\%} & \textbf{99.27\%} & \textbf{99.20\%} & \textbf{99.45\%} & \textbf{99.30\%} \\
\hline
\end{tabular}
}
\end{center}
\label{tab:pixel_acc}
\vspace{-0.5cm}
\end{table}

% \begin{table*}[!t]
% \caption{Pixel Accuracy of \textit{SIO-Net}}
% \begin{center}
% \resizebox{\linewidth}{!}{
% \begin{tabular}{|c|cccc|cccc|cccc|cccc|}
% \hline
% \multirow{2}{*}{\textbf{Method}} & \multicolumn{4}{c|}{Pangyo} & \multicolumn{4}{c|}{Sangam} & \multicolumn{4}{c|}{Yeouido} & \multicolumn{4}{c|}{Magok} \\
% & $W,B$ & $W,L$ & $Y$ & Total & $W,B$ & $W,L$ & $Y$ & Total & $W,B$ & $W,L$ & $Y$ & Total & $W,B$ & $W,L$ & $Y$ & Total \\
% \hline
% \textbf{HDMapNet} & - & - & - & 87.61\% & - & - & - & 91.85\% & - & - & - & - & - & - & - & 44.68\% \\
% \textbf{\textit{SIO-Net}} & 99.23\% & 99.85\% & 98.68\% & \textbf{99.26\%} & 99.32\% & 99.88\% & 98.60\% & \textbf{99.27\%} & 99.38\% & 99.89\% & 98.34\% & \textbf{99.20\%} & 99.58\% & 99.94\% & 98.83\% & \textbf{99.45\%} \\
% \hline
% \end{tabular}
% }
% \end{center}
% \label{tab:pixel_acc}
% \vspace{-0.5cm}
% \end{table*}
\begin{table}[!t]
\caption{mIoU of \textit{SIO-Net}}
\begin{center}
\resizebox{\linewidth}{!}{
\begin{tabular}{|c|c|cccc|c|}
\hline
\multirow{2}{*}{\textbf{Network}}          & \multirow{2}{*}{\textbf{Lane Type}} & \multirow{2}{*}{Pangyo} & \multirow{2}{*}{Sangam} & \multirow{2}{*}{Yeouido} & \multirow{2}{*}{Magok} & \multirow{2}{*}{\textbf{Average}} \\
&&&&&&\\
\hline
\multirow{2}{*}{\textbf{HDMapNet}}         & \multirow{2}{*}{\textbf{Total}}         & \multirow{2}{*}{\textbf{0.8323}} & \multirow{2}{*}{\textbf{0.8788}} & \multirow{2}{*}{-} & \multirow{2}{*}{\textbf{0.4625}} & \multirow{2}{*}{\textbf{0.7245}} \\
&&&&&&\\
\hline
\multirow{4}{*}{\textbf{\textit{SIO-Net}}} & White, Broken  & 0.5533 & 0.5713 & 0.5147          & 0.4530 & 0.5231 \\
                                           & White          & 0.2542 & 0.3141 & 0.2357          & 0.1703 & 0.2436 \\
                                           & Yellow         & 0.4429 & 0.5031 & 0.4652          & 0.5023 & 0.4784 \\
                                           & \textbf{Total} & 0.4168 & 0.4628 & \textbf{0.4052} & 0.3752 & 0.4150 \\
\hline
\multirow{2}{*}{\textbf{Lane Ratio}}       & \multirow{2}{*}{\textbf{Total}} & \multirow{2}{*}{\textbf{1.13\%}} & \multirow{2}{*}{\textbf{1.34\%}} & \multirow{2}{*}{\textbf{1.31\%}} & \multirow{2}{*}{\textbf{0.95\%}} & \multirow{2}{*}{\textbf{1.18\%}} \\
&&&&&&\\
\hline
\end{tabular}
}
\end{center}
\label{tab:miou}
\vspace{-0.5cm}
\end{table}

% \begin{table*}[!t]
% \caption{mIoU of \textit{SIO-Net}}
% \begin{center}
% \resizebox{\linewidth}{!}{
% \begin{tabular}{|c|cccc|cccc|cccc|cccc|}
% \hline
% \multirow{2}{*}{\textbf{Method}} & \multicolumn{4}{c|}{Pangyo} & \multicolumn{4}{c|}{Sangam} & \multicolumn{4}{c|}{Yeouido} & \multicolumn{4}{c|}{Magok} \\
% & $W,B$ & $W,L$ & $Y$ & Total & $W,B$ & $W,L$ & $Y$ & Total & $W,B$ & $W,L$ & $Y$ & Total & $W,B$ & $W,L$ & $Y$ & Total \\
% \hline
% \textbf{HDMapNet} & - & - & - & \textbf{0.8323} & - & - & - & \textbf{0.8788} & - & - & - & - & - & - & - & \textbf{0.4625} \\
% \textbf{\textit{SIO-Net}} & 0.5533 & 0.2542 & 0.4429 & 0.4168 & 0.5713 & 0.3141 & 0.5031 & 0.4628 & 0.5147 & 0.2357 & 0.4652 & \textbf{0.4052} & 0.4530 & 0.1703 & \textbf{0.5023} & 0.3752 \\
% \hline
% \textbf{Lane Ratio} & 1.52\% & 0.21\% & 1.66\% & 1.13\% & 1.66\% & 0.20\% & 2.15\% & 1.34\% & 1.34\% & 0.19\% & 2.40\% & 1.31\% & 1.05\% & 0.08\% & 1.74\% & 0.95\% \\
% \hline
% \end{tabular}
% }
% \end{center}
% \label{tab:miou}
% \vspace{-0.5cm}
% \end{table*}
\begin{table}[!t]
\caption{Evaluation of Global HD Lane Map}
\begin{center}
\resizebox{\linewidth}{!}{
\begin{tabular}{|c|c|cccc|c|}
\hline
\multirow{2}{*}{\textbf{Network}}          & \multirow{2}{*}{\textbf{Metric}} & \multirow{2}{*}{Pangyo} & \multirow{2}{*}{Sangam} & \multirow{2}{*}{Yeouido} & \multirow{2}{*}{Magok} & \multirow{2}{*}{\textbf{Average}} \\
&&&&&&\\
\hline
\multirow{5}{*}{\textbf{HDMapNet}} & $C_m$      & 40.76\%   & 30.27\%   & - & 49.49\%   & 40.17\% \\
                                   & $A_m(0.25)$ & 46.58\%   & 11.73\%   & - & 41.08\%  & 33.13\% \\
                                   & $A_m(1.0)$ & 67.85\%   & 36.90\%   & - & 65.32\%   & 56.69\% \\
                                   & $A_m(1.5)$ & 76.46\%   & 48.81\%   & - & 72.05\%   & 65.77\% \\
                                   & $D_v$      & 0.3856\,m & 1.5724\,m & - & 0.4503\,m & 0.8028\,m \\
\hline
\multirow{5}{*}{\textbf{\textit{SIO-Net}}} & $C_m$      & \textbf{63.65\%}   & \textbf{67.33\%}   & \textbf{63.20\%}   & \textbf{73.18\%} & \textbf{66.84\%} \\
                                       & $A_m(0.25)$ & \textbf{54.11\%}   & \textbf{18.71\%}   & \textbf{20.25\%}   & \textbf{47.23\%} & \textbf{35.08\%} \\
                                       & $A_m(1.0)$ & \textbf{70.47\%}   & \textbf{52.08\%}   & \textbf{50.46\%}   & \textbf{75.74\%} & \textbf{62.19\%} \\
                                       & $A_m(1.5)$ & \textbf{77.01\%}   & \textbf{66.52\%}   & \textbf{60.93\%}   & \textbf{82.94\%} & \textbf{71.85\%} \\
                                       & $D_v$      & \textbf{0.2405\,m} & \textbf{0.9416\,m} & \textbf{1.0463\,m} & \textbf{0.3568\,m} & \textbf{0.6463\,m} \\
\hline
\end{tabular}
}
\end{center}
\label{tab:global}
\vspace{-0.5cm}
\end{table}
\begin{table}[!t]
\caption{Evaluation using Ground Truth Lane Images}
\begin{center}
\resizebox{\linewidth}{!}{
\begin{tabular}{|c|c|cccc|c|}
\hline
\multirow{2}{*}{\textbf{Network}}          & \multirow{2}{*}{\textbf{Metric}} & \multirow{2}{*}{Pangyo} & \multirow{2}{*}{Sangam} & \multirow{2}{*}{Yeouido} & \multirow{2}{*}{Magok} & \multirow{2}{*}{\textbf{Average}} \\
&&&&&&\\
\hline
\multirow{5}{*}{\textbf{Naver Labs}} & $C_m$       & 83.07\%   & 79.20\%   & 75.21\%   & 86.97\%   & 81.11\% \\
                                     & $A_m(0.25)$ & 84.89\%   & 64.06\%   & 67.29\%   & 85.71\%   & 75.49\% \\
                                     & $A_m(1.0)$  & 91.90\%   & 82.86\%   & 85.81\%   & 94.26\%   & 88.71\% \\
                                     & $A_m(1.5)$  & 94.07\%   & 88.82\%   & 89.92\%   & 95.73\%   & 92.14\% \\
                                     & $D_v$       & 0.0588\,m & 0.1579\,m & 0.1563\,m & 0.0666\,m & 0.1099\,m \\
\hline
\multirow{2}{*}{} & \multirow{2}{*}{} & \multirow{2}{*}{Boston} & \multirow{2}{*}{North} & \multirow{2}{*}{Holland} & \multirow{2}{*}{Queens} & \multirow{2}{*}{} \\
&&&&&&\\
\hline

\multirow{5}{*}{\textbf{NuScenes}} & $C_m$       & 92.08\%   & 98.62\%   & 92.66\%   & 89.74\%   & 93.28\% \\
                                   & $A_m(0.25)$ & 90.36\%   & 97.45\%   & 87.16\%   & 85.08\%   & 90.01\% \\
                                   & $A_m(1.0)$  & 93.42\%   & 99.41\%   & 90.21\%   & 87.18\%   & 92.56\% \\
                                   & $A_m(1.5)$  & 94.27\%   & 99.41\%   & 90.52\%   & 87.65\%   & 92.96\% \\
                                   & $D_v$       & 0.0743\,m & 0.0455\,m & 0.0659\,m & 0.0649\,m & 0.0627\,m \\
\hline
\end{tabular}
}
\end{center}
\label{tab:gt}
\vspace{-0.5cm}
\end{table}

\begin{figure}[!t]
\centerline{\includegraphics[width=0.98\linewidth]{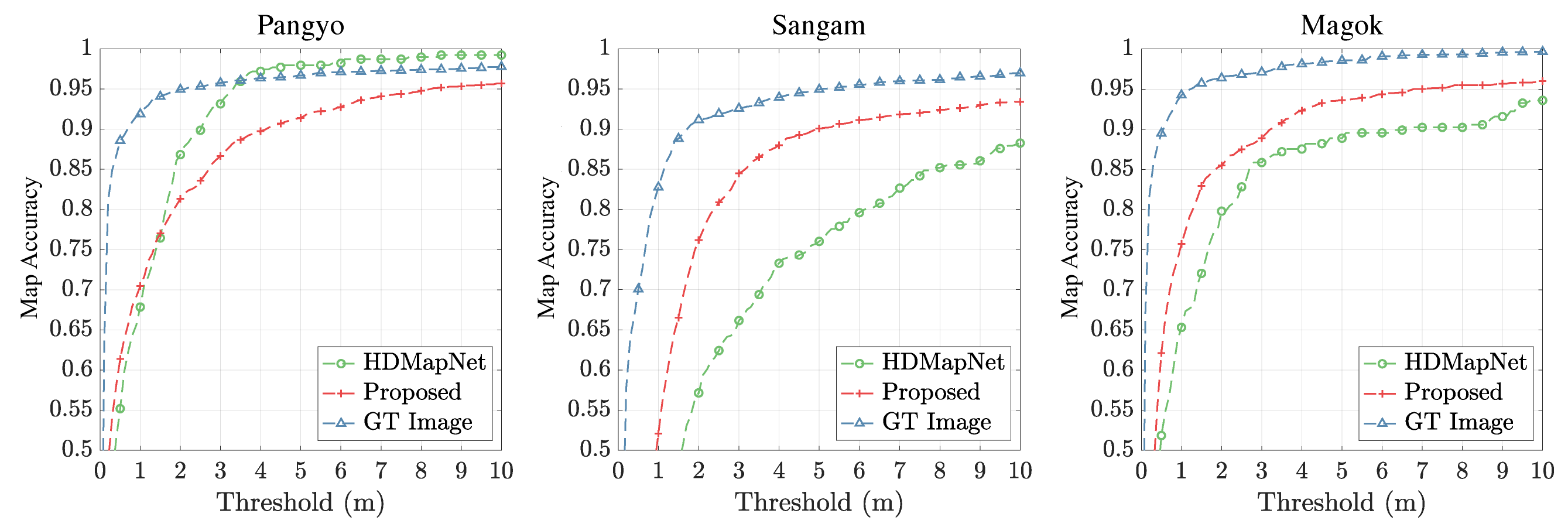}}
\caption{Map accuracy, $A_m$, based on different thresholds for each sequence. Since the NAVER LABS Open Dataset does not provide LiDAR and camera data for the Yeouido sequence, there are no results using HDmapNet for the Yeouido sequence.}
\vspace{-0.5cm}
\label{fig:acc_threshold}
\end{figure}

\subsection{SIO-Mapper Evaluation}
Finally, we evaluate the lane-level \ac{HD} map accuracy on four sequences of Naver Labs Open Dataset. We chose three threshold of map accuracy, $A_m$: 0.25\,m, 1.0\,m, and 1.5\,m which represent the pixel resolution of the satellite image, half of the average vehicle size, and half of the average lane width, respectively. Depending on the purpose of the lane-level \ac{HD} map, such as path planning and localization, users can choose different threshold values for suitable evaluation.

\tabref{tab:global} details the quantitative evaluation of resulting lane-level \ac{HD} maps in four sequences. As shown in the table, SIO-Net considerably outperforms HDMapNet even SIO-Net only utilizes \ac{OSM} and satellite images while HDMapNet utilizes a LiDAR scan and six surrounding images. Also, suggested method shows mean vertex distance, $D_v$, 0.6463/,m in average which is approximately $1/5$ of the average lane width. However, as shown in \figref{fig:acc_threshold}, HDMapNet performs better when threshold is larger than 2\,m in Pangyo sequence. This is occured because SIO-Net totally failed in some road in Pangyo sequence.

\tabref{tab:gt} shows the evaluation lane-level \ac{HD} maps using ground truth lane images in eight sequences from two datasets. We compared it with the result using lane images generated from SIO-Net and HDMapNet to validate SIO-Net generates more proper lane images for the lane-level \ac{HD} map construction. As the table shows, we guarantee over 83\% of lanes generated with at least 64\%, 82\%, and 87\% accuracy for each threshold using proposed framework. Also, we guarantee meter-level accuracy in terms of distance.

\begin{figure}[!t]
\centering
\subfigure[Pangyo, overall view]{\includegraphics[width=0.48\linewidth]{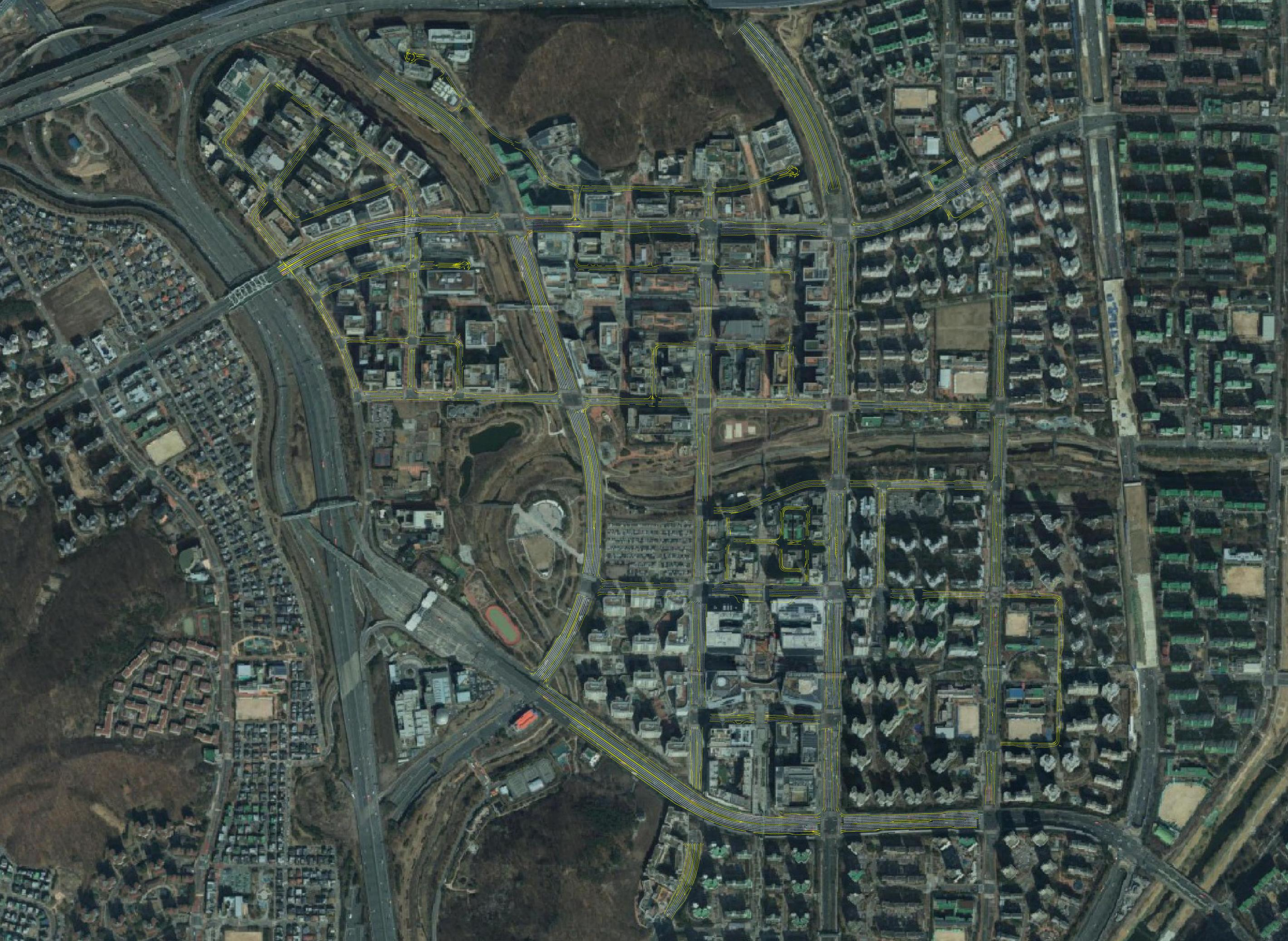}}
\subfigure[Sangam, overall view]{\includegraphics[width=0.48\linewidth]{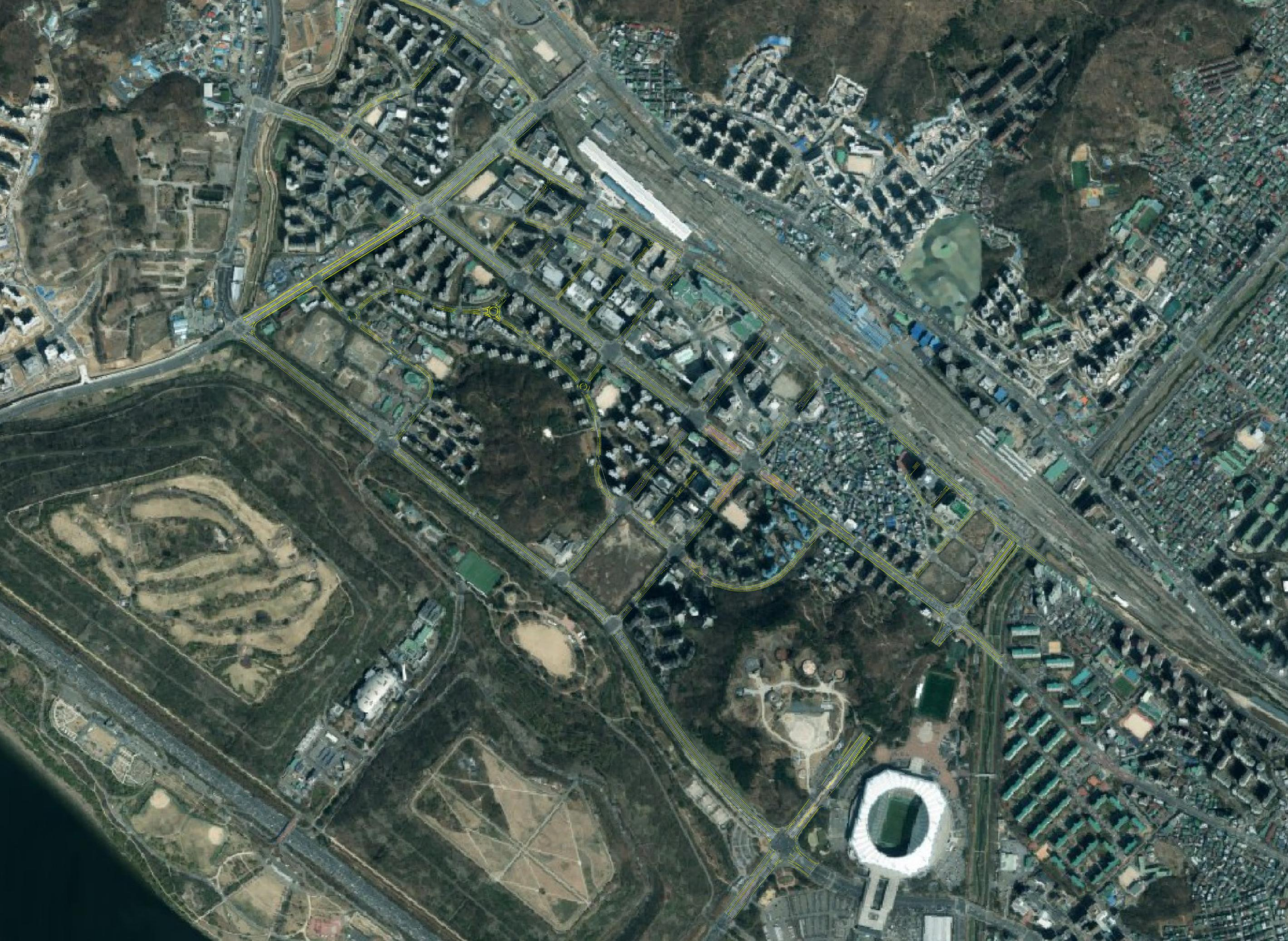}} \\
\subfigure[Yeouido, overall view]{\includegraphics[width=0.48\linewidth]{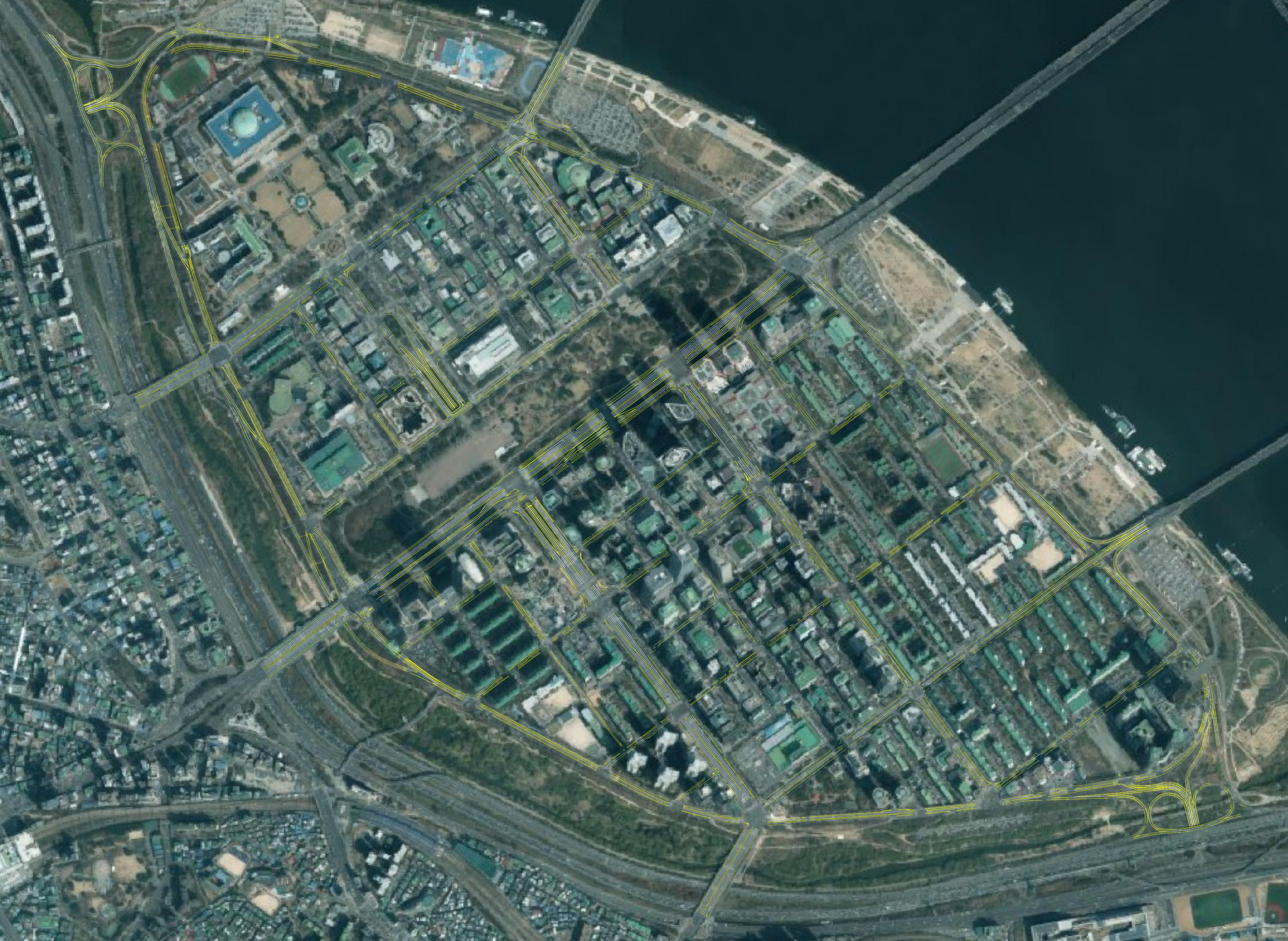}}
\subfigure[Magok, overall view]{\includegraphics[width=0.48\linewidth]{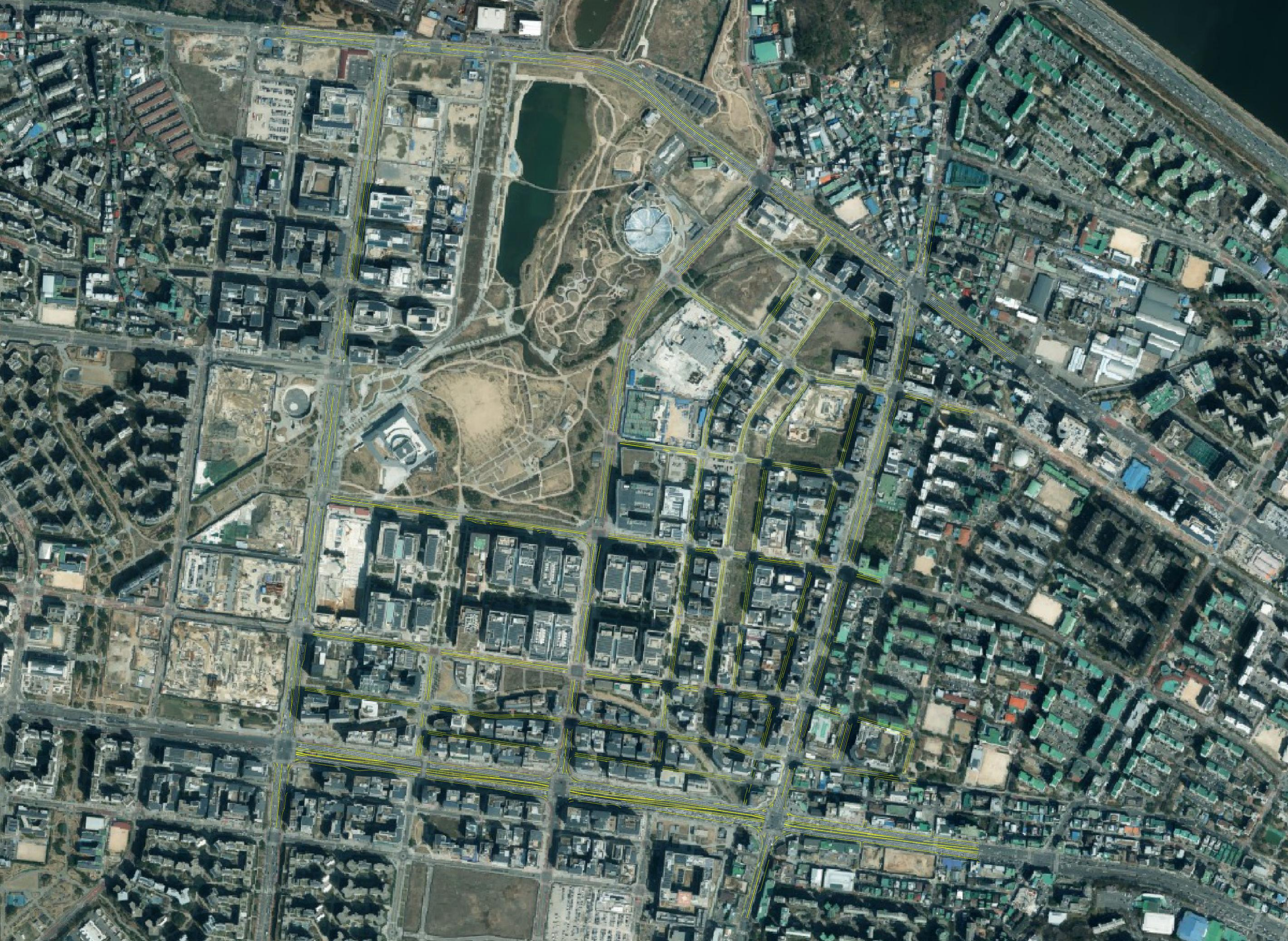}}
\caption{Resulting city-scale lane-level HD maps for each sequences.}
\label{fig:result_map}
% \vspace{0.5cm}
\end{figure}
\begin{figure}[!t]
\centering
\subfigure[U-turn lane]{\includegraphics[width=0.48\linewidth, height=0.3\linewidth]{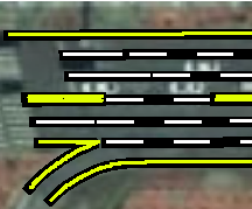}}
\subfigure[Narrow merging roads]{\includegraphics[width=0.48\linewidth, height=0.3\linewidth]{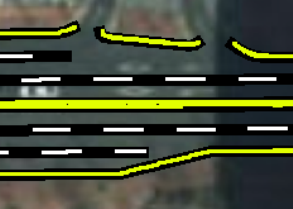}} \\
\subfigure[Painted island (Stop line)]{\includegraphics[width=0.48\linewidth, height=0.3\linewidth]{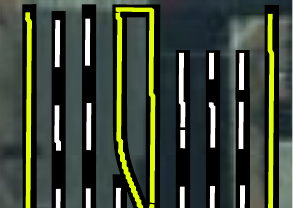}}
\subfigure[Painted island]{\includegraphics[width=0.48\linewidth, height=0.3\linewidth]{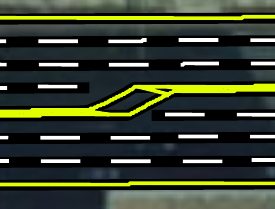}} \\
\subfigure[Roundabout]{\includegraphics[width=0.48\linewidth, height=0.3\linewidth]{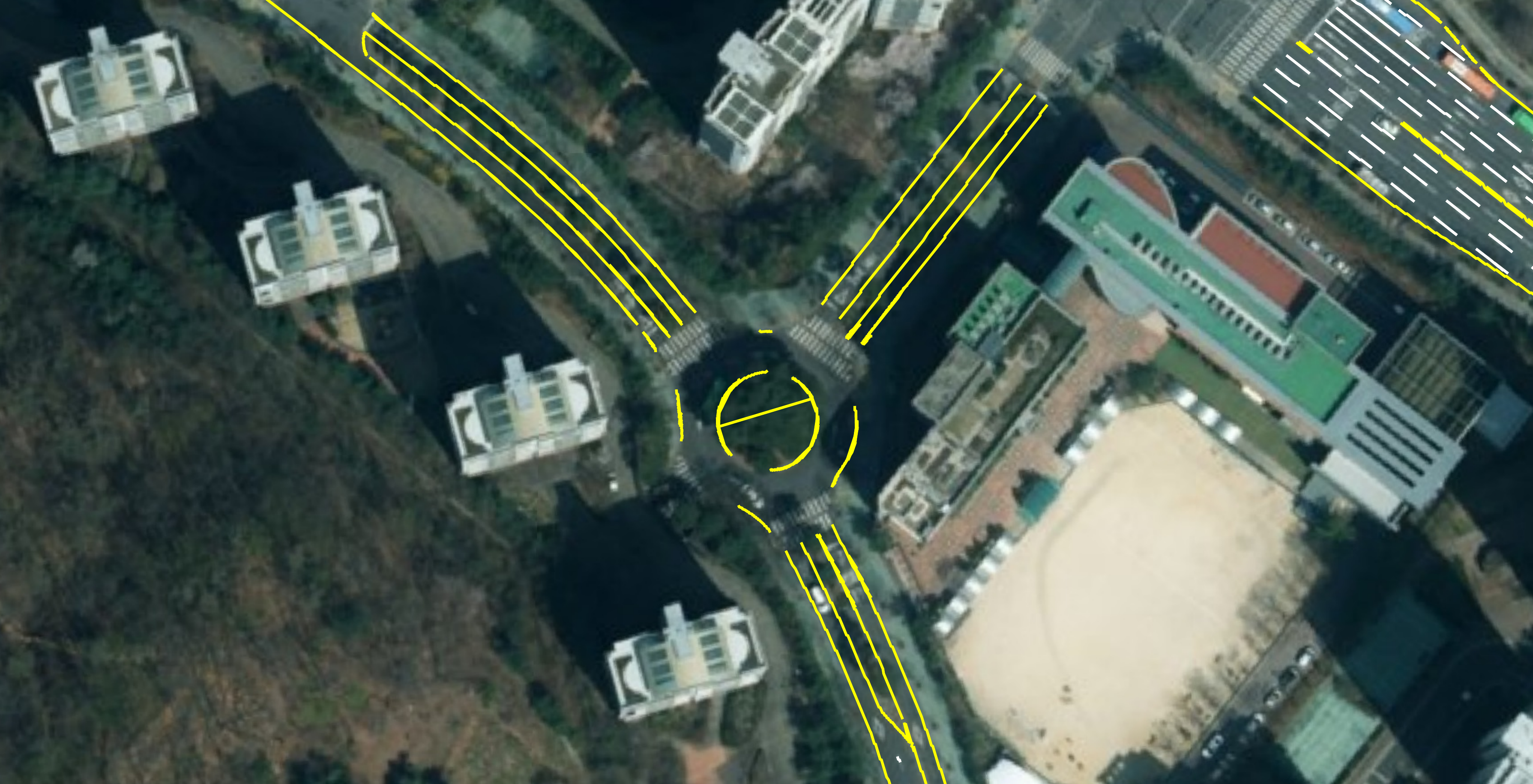}}
\subfigure[Under Overpass + Shadow]{\includegraphics[width=0.48\linewidth, height=0.3\linewidth]{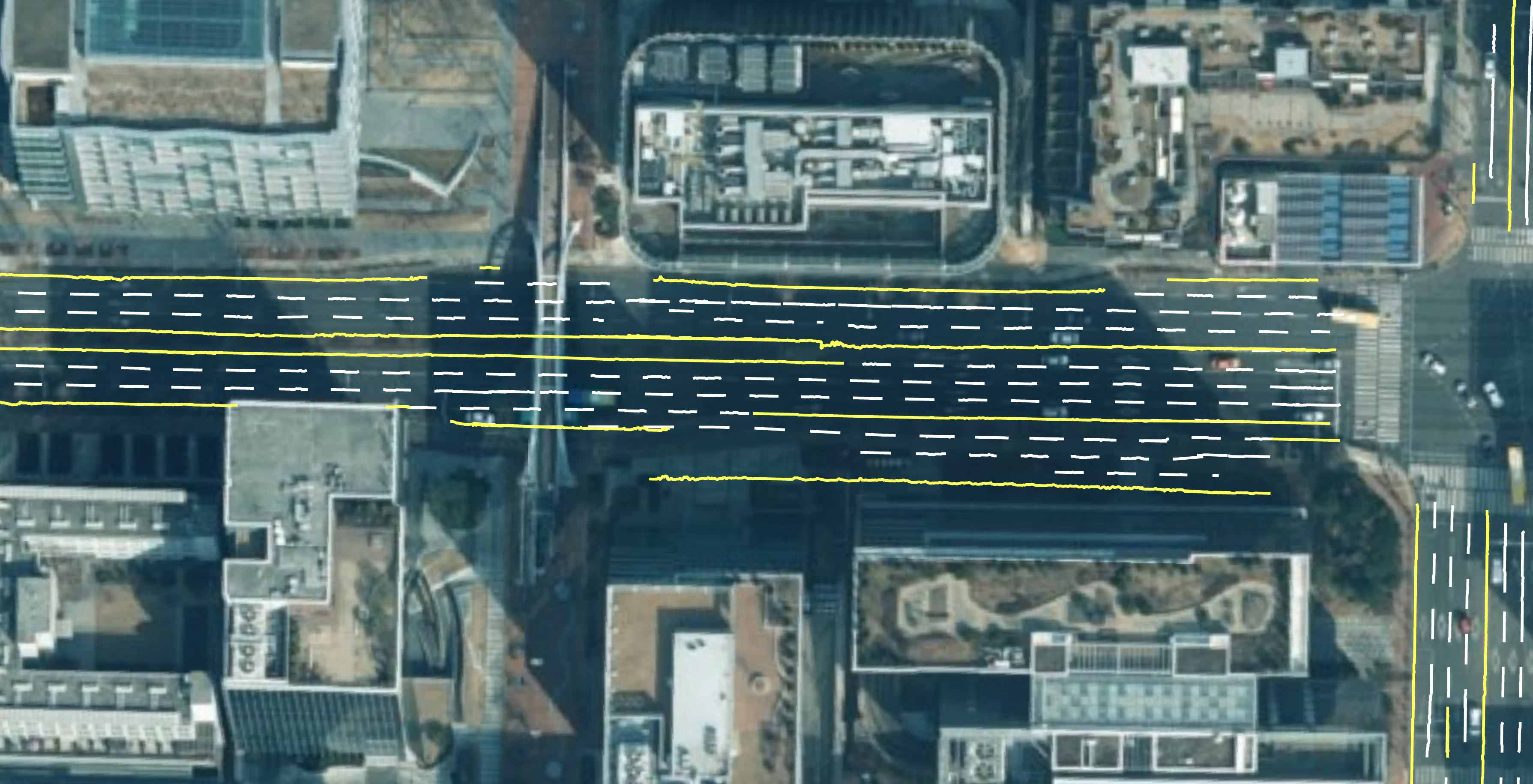}}
\caption{Resulting lane-level HD map in various road environments. Black represents ground truth and colored lanes represent resulting lanes. (a): An example at the U-turn lane, where yellow and white lanes are mixed in one straight line. (b): An example at narrow merging roads with small yellow line fragments. (c), (d): Painted islands at the end of the road and middle of the road. (e): Roundabout with curved lanes. (f): Under overpass in urban environment with shadowed condition.}
\label{fig:qualitative}
% \vspace{0.5cm}
\end{figure}

Also, the qualitative lane-level \ac{HD} map construction result shows that our suggested framework SIO-Mapper constructs city-scale lane-level \ac{HD} maps in various environments as shown in \figref{fig:result_map}. Moreover, as shown in \figref{fig:qualitative}, SIO-Mapper constructs lane-level \ac{HD} maps in challenging environments such as U-turn lanes, merging roads, painted islands, roundabouts, and overpasses.
\section{Discussion and Conclusion}
In this paper, we successfully constructed city-scale lane-level \ac{HD} maps, overcoming key limitations of previous approaches. SIO-Mapper framework not only eliminates the need of costly and time-consuming sensor-based physical visits but also achieves accurate lane-level \ac{HD} maps in wide areas. To the best of our knowledge, this is the first fully automated approach that constructs a lane-level \ac{HD} map from publicly available satellite images and OpenStreetMap road information, without requiring direct sensor measurements. Also, one of the major contributions of this paper is suggesting unified evaluation metrics specifically designed for vectorized lane-level \ac{HD} maps. By introducing three new metrics, map coverage, map accuracy, and mean vertex distance, we provide comprehensive evaluation for both accuracy and coverage of lane-level \ac{HD} maps. SIO-Mapper was validated across eight sequences from two datasets, demonstrating robust performance in diverse road environments with 87.20\% map coverage and 71.85\% map accuracy at 1.5\,m threshold on average.

However, there are still some limitations to address. SIO-Mapper heavily relies on the quality of the OpenStreetMap data and satellite images. Since OpenStreetMap is a crowd-sourced data, it may contain inaccuracies or outdated information, potentially affecting the resulting maps. Additionally, while satellite images are generally reliable, they are not updated frequently due to the high cost of satellite image acquisition. As a result, there can be mismatches between the satellite images and real-time road conditions. Future research should explore change detection and frequent updating of satellite images to ensure that the constructed \ac{HD} maps remain accurate and up to date.

\addcontentsline{toc}{chapter}{Bibliography}
\renewcommand*{\bibfont}{\small}
\bibliographystyle{IEEEtranN} % not IEEEtran, but IEEEtranN for using citeauthor
\bibliography{reference}

% Generated by IEEEtranN.bst, version: 1.14 (2015/08/26)
\begin{thebibliography}{27}
\providecommand{\natexlab}[1]{#1}
\providecommand{\url}[1]{#1}
\csname url@samestyle\endcsname
\providecommand{\newblock}{\relax}
\providecommand{\bibinfo}[2]{#2}
\providecommand{\BIBentrySTDinterwordspacing}{\spaceskip=0pt\relax}
\providecommand{\BIBentryALTinterwordstretchfactor}{4}
\providecommand{\BIBentryALTinterwordspacing}{\spaceskip=\fontdimen2\font plus
\BIBentryALTinterwordstretchfactor\fontdimen3\font minus \fontdimen4\font\relax}
\providecommand{\BIBforeignlanguage}[2]{{%
\expandafter\ifx\csname l@#1\endcsname\relax
\typeout{** WARNING: IEEEtranN.bst: No hyphenation pattern has been}%
\typeout{** loaded for the language `#1'. Using the pattern for}%
\typeout{** the default language instead.}%
\else
\language=\csname l@#1\endcsname
\fi
#2}}
\providecommand{\BIBdecl}{\relax}
\BIBdecl

\bibitem[nav()]{naverlabs}
``Naver labs open dataset: Hd map \& localization dataset,'' \url{https://www.naverlabs.com/datasets}, accessed: 2023-04-19.

\bibitem[Caesar et~al.(2020)Caesar, Bankiti, Lang, Vora, Liong, Xu, Krishnan, Pan, Baldan, and Beijbom]{nuscenes}
H.~Caesar, V.~Bankiti, A.~H. Lang, S.~Vora, V.~E. Liong, Q.~Xu, A.~Krishnan, Y.~Pan, G.~Baldan, and O.~Beijbom, ``nuscenes: A multimodal dataset for autonomous driving,'' in \emph{CVPR}, 2020.

\bibitem[Guo et~al.(2014)Guo, Meguro, Kojima, and Naito]{guo2014automatic}
C.~Guo, J.-i. Meguro, Y.~Kojima, and T.~Naito, ``Automatic lane-level map generation for advanced driver assistance systems using low-cost sensors,'' in \emph{2014 IEEE international conference on robotics and automation (ICRA)}.\hskip 1em plus 0.5em minus 0.4em\relax IEEE, 2014, pp. 3975--3982.

\bibitem[Guo et~al.(2016)Guo, Kidono, Meguro, Kojima, Ogawa, and Naito]{guo2016low}
C.~Guo, K.~Kidono, J.~Meguro, Y.~Kojima, M.~Ogawa, and T.~Naito, ``A low-cost solution for automatic lane-level map generation using conventional in-car sensors,'' \emph{IEEE Transactions on Intelligent Transportation Systems}, vol.~17, no.~8, pp. 2355--2366, 2016.

\bibitem[Gwon et~al.(2016)Gwon, Hur, Kim, and Seo]{gwon2016generation}
G.-P. Gwon, W.-S. Hur, S.-W. Kim, and S.-W. Seo, ``Generation of a precise and efficient lane-level road map for intelligent vehicle systems,'' \emph{IEEE Transactions on Vehicular Technology}, vol.~66, no.~6, pp. 4517--4533, 2016.

\bibitem[Joshi and James(2015)]{joshi2015generation}
A.~Joshi and M.~R. James, ``Generation of accurate lane-level maps from coarse prior maps and lidar,'' \emph{IEEE Intelligent Transportation Systems Magazine}, vol.~7, no.~1, pp. 19--29, 2015.

\bibitem[Liang et~al.(2019)Liang, Homayounfar, Ma, Wang, and Urtasun]{liang2019convolutional}
J.~Liang, N.~Homayounfar, W.-C. Ma, S.~Wang, and R.~Urtasun, ``Convolutional recurrent network for road boundary extraction,'' in \emph{Proceedings of the IEEE/CVF Conference on Computer Vision and Pattern Recognition}, 2019, pp. 9512--9521.

\bibitem[Homayounfar et~al.(2019)Homayounfar, Ma, Liang, Wu, Fan, and Urtasun]{homayounfar2019dagmapper}
N.~Homayounfar, W.-C. Ma, J.~Liang, X.~Wu, J.~Fan, and R.~Urtasun, ``Dagmapper: Learning to map by discovering lane topology,'' in \emph{Proceedings of the IEEE/CVF International Conference on Computer Vision}, 2019, pp. 2911--2920.

\bibitem[Z{\"u}rn et~al.(2021)Z{\"u}rn, Vertens, and Burgard]{zurn2021lane}
J.~Z{\"u}rn, J.~Vertens, and W.~Burgard, ``Lane graph estimation for scene understanding in urban driving,'' \emph{IEEE Robotics and Automation Letters}, vol.~6, no.~4, pp. 8615--8622, 2021.

\bibitem[Yang et~al.(2018)Yang, Lu, Lee, Batra, and Parikh]{yang2018graph}
J.~Yang, J.~Lu, S.~Lee, D.~Batra, and D.~Parikh, ``Graph r-cnn for scene graph generation,'' in \emph{Proceedings of the European conference on computer vision (ECCV)}, 2018, pp. 670--685.

\bibitem[Zhou et~al.(2021)Zhou, Takeda, Tomizuka, and Zhan]{zhou2021automatic}
Y.~Zhou, Y.~Takeda, M.~Tomizuka, and W.~Zhan, ``Automatic construction of lane-level hd maps for urban scenes,'' in \emph{2021 IEEE/RSJ International Conference on Intelligent Robots and Systems (IROS)}.\hskip 1em plus 0.5em minus 0.4em\relax IEEE, 2021, pp. 6649--6656.

\bibitem[Chen et~al.(2018)Chen, Zhu, Papandreou, Schroff, and Adam]{chen2018encoder}
L.-C. Chen, Y.~Zhu, G.~Papandreou, F.~Schroff, and H.~Adam, ``Encoder-decoder with atrous separable convolution for semantic image segmentation,'' in \emph{Proceedings of the European conference on computer vision (ECCV)}, 2018, pp. 801--818.

\bibitem[Li et~al.(2022)Li, Wang, Wang, and Zhao]{li2022hdmapnet}
Q.~Li, Y.~Wang, Y.~Wang, and H.~Zhao, ``Hdmapnet: An online hd map construction and evaluation framework,'' in \emph{2022 International Conference on Robotics and Automation (ICRA)}.\hskip 1em plus 0.5em minus 0.4em\relax IEEE, 2022, pp. 4628--4634.

\bibitem[Liu et~al.(2023)Liu, Guan, Yuan, Liu, Zhou, Kun, Li, Zheng, and Mei]{liu2023learning}
R.~Liu, Z.~Guan, Z.~Yuan, A.~Liu, T.~Zhou, T.~Kun, E.~Li, C.~Zheng, and S.~Mei, ``Learning to detect 3d lanes by shape matching and embedding,'' in \emph{Proceedings of the IEEE/CVF Winter Conference on Applications of Computer Vision}, 2023, pp. 4291--4299.

\bibitem[Ort et~al.(2022)Ort, Walls, Parkison, Gilitschenski, and Rus]{ort2022maplite}
T.~Ort, J.~M. Walls, S.~A. Parkison, I.~Gilitschenski, and D.~Rus, ``Maplite 2.0: Online hd map inference using a prior sd map,'' \emph{IEEE Robotics and Automation Letters}, vol.~7, no.~3, pp. 8355--8362, 2022.

\bibitem[M{\'a}ttyus et~al.(2017)M{\'a}ttyus, Luo, and Urtasun]{mattyus2017deeproadmapper}
G.~M{\'a}ttyus, W.~Luo, and R.~Urtasun, ``Deeproadmapper: Extracting road topology from aerial images,'' in \emph{Proceedings of the IEEE international conference on computer vision}, 2017, pp. 3438--3446.

\bibitem[He et~al.(2016)He, Zhang, Ren, and Sun]{he2016deep}
K.~He, X.~Zhang, S.~Ren, and J.~Sun, ``Deep residual learning for image recognition,'' in \emph{Proceedings of the IEEE conference on computer vision and pattern recognition}, 2016, pp. 770--778.

\bibitem[Azimi et~al.(2018)Azimi, Fischercs, K{\"o}rner, and Reinartz]{azimi2018aerial}
S.~M. Azimi, P.~Fischercs, M.~K{\"o}rner, and P.~Reinartz, ``Aerial lanenet: Lane-marking semantic segmentation in aerial imagery using wavelet-enhanced cost-sensitive symmetric fully convolutional neural networks,'' \emph{IEEE Transactions on Geoscience and Remote Sensing}, vol.~57, no.~5, pp. 2920--2938, 2018.

\bibitem[He and Balakrishnan(2022)]{he2022lane}
S.~He and H.~Balakrishnan, ``Lane-level street map extraction from aerial imagery,'' in \emph{Proceedings of the IEEE/CVF Winter Conference on Applications of Computer Vision}, 2022, pp. 2080--2089.

\bibitem[Ronneberger et~al.(2015)Ronneberger, Fischer, and Brox]{ronneberger2015u}
O.~Ronneberger, P.~Fischer, and T.~Brox, ``U-net: Convolutional networks for biomedical image segmentation,'' in \emph{Medical Image Computing and Computer-Assisted Intervention--MICCAI 2015: 18th International Conference, Munich, Germany, October 5-9, 2015, Proceedings, Part III 18}.\hskip 1em plus 0.5em minus 0.4em\relax Springer, 2015, pp. 234--241.

\bibitem[Xu et~al.(2021{\natexlab{a}})Xu, Sun, Wang, and Liu]{xu2021cp}
Z.~Xu, Y.~Sun, L.~Wang, and M.~Liu, ``Cp-loss: Connectivity-preserving loss for road curb detection in autonomous driving with aerial images,'' in \emph{2021 IEEE/RSJ International Conference on Intelligent Robots and Systems (IROS)}.\hskip 1em plus 0.5em minus 0.4em\relax IEEE, 2021, pp. 1117--1123.

\bibitem[Xu et~al.(2021{\natexlab{b}})Xu, Sun, and Liu]{xu2021icurb}
Z.~Xu, Y.~Sun, and M.~Liu, ``icurb: Imitation learning-based detection of road curbs using aerial images for autonomous driving,'' \emph{IEEE Robotics and Automation Letters}, vol.~6, no.~2, pp. 1097--1104, 2021.

\bibitem[Lin et~al.(2017)Lin, Doll{\'a}r, Girshick, He, Hariharan, and Belongie]{lin2017feature}
T.-Y. Lin, P.~Doll{\'a}r, R.~Girshick, K.~He, B.~Hariharan, and S.~Belongie, ``Feature pyramid networks for object detection,'' in \emph{Proceedings of the IEEE conference on computer vision and pattern recognition}, 2017, pp. 2117--2125.

\bibitem[Xu et~al.(2022{\natexlab{a}})Xu, Liu, Gan, Hu, Sun, Liu, and Wang]{xu2022csboundary}
Z.~Xu, Y.~Liu, L.~Gan, X.~Hu, Y.~Sun, M.~Liu, and L.~Wang, ``csboundary: City-scale road-boundary detection in aerial images for high-definition maps,'' \emph{IEEE Robotics and Automation Letters}, vol.~7, no.~2, pp. 5063--5070, 2022.

\bibitem[Xu et~al.(2022{\natexlab{b}})Xu, Liu, Gan, Sun, Wu, Liu, and Wang]{xu2022rngdet}
Z.~Xu, Y.~Liu, L.~Gan, Y.~Sun, X.~Wu, M.~Liu, and L.~Wang, ``Rngdet: Road network graph detection by transformer in aerial images,'' \emph{IEEE Transactions on Geoscience and Remote Sensing}, vol.~60, pp. 1--12, 2022.

\bibitem[Vaswani et~al.(2017)Vaswani, Shazeer, Parmar, Uszkoreit, Jones, Gomez, Kaiser, and Polosukhin]{vaswani2017attention}
A.~Vaswani, N.~Shazeer, N.~Parmar, J.~Uszkoreit, L.~Jones, A.~N. Gomez, {\L}.~Kaiser, and I.~Polosukhin, ``Attention is all you need,'' \emph{Advances in neural information processing systems}, vol.~30, 2017.

\bibitem[Ester et~al.(1996)Ester, Kriegel, Sander, Xu, et~al.]{ester1996density}
M.~Ester, H.-P. Kriegel, J.~Sander, X.~Xu \emph{et~al.}, ``A density-based algorithm for discovering clusters in large spatial databases with noise.'' in \emph{kdd}, vol.~96, no.~34, 1996, pp. 226--231.

\end{thebibliography}
\section{Biography} 
\begin{IEEEbiography}[{\includegraphics[width=1in,height=1.25in,clip,keepaspectratio]{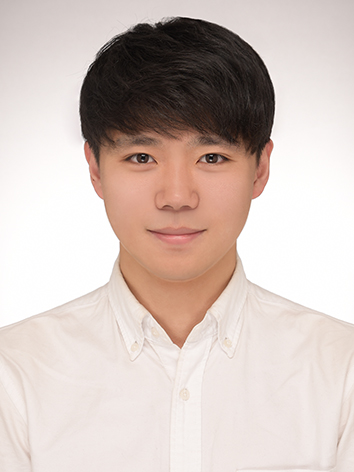}}]{Younghun Cho}
received the B.S. degree in civil and environmental engineering from KAIST, Daejeon, Korea, in 2018, the M.S. degree in civil and environmental engineering from KAIST, Daejeon, Korea, in 2020, and the Ph.D. degree in civil and environmental engineering from KAIST, Daejeon, Korea, in 2024. Currently, he is an post doctoral researcher at Korea Institute of Machinery and Materials. His research interests include High-Definition mapping and LiDAR localization.
\end{IEEEbiography}

\vspace{-15cm}

\begin{IEEEbiography}[{\includegraphics[width=1in,height=1.25in,clip,keepaspectratio]{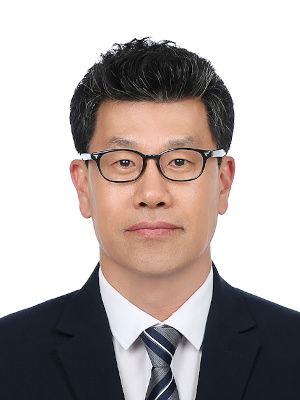}}]{Jee-Hwan Ryu}
(Senior Member, IEEE) received the B.S. degree from Inha University, Incheon, South Korea, in 1995, and the M.S. and Ph.D. degrees from the Korea Advanced Institute of Science and Technology, Daejeon, South Korea, in 1997 and 2002, respectively, all in mechanical engineering. He is currently a Professor with the Department of Civil and Environmental Engineering, Korea Advanced Institute of Science and Technology. His research interests include haptics, telerobotics, exoskeletons, soft robots, and autonomous vehicles.
\end{IEEEbiography}

\end{document}